\pdfoutput=1

\documentclass[10pt,twocolumn,letterpaper]{article}

\usepackage{cvpr}              

\usepackage[accsupp]{axessibility}  

\definecolor{cvprblue}{rgb}{0.21,0.49,0.74}
\usepackage[pagebackref,breaklinks,colorlinks,allcolors=cvprblue]{hyperref}

\def\paperID{6347} 
\def\confName{CVPR}
\def\confYear{2025}

\title{Sim-to-Real Causal Transfer:\\A Metric Learning Approach to Causally-Aware Interaction Representations}

\author{
Ahmad Rahimi$^{{1},}\thanks{Equal contribution.}$ \quad
Po-Chien Luan$^{{1},}\footnotemark[1]$ \quad
Yuejiang Liu$^{{2},}\footnotemark[1]$ \quad
Frano Rajic$^3$ \quad
Alexandre Alahi$^1$ \\
$^1$ École Polytechnique Fédérale de Lausanne (EPFL), $^2$ Stanford University, $^3$ ETH Zurich \\
{\tt\small \{ahmad.rahimi, po-chien.luan, alexandre.alahi\}@epfl.ch}
}

\usepackage{amsmath,amsfonts,bm}

\def\eqref#1{equation~\ref{#1}}

\def\1{\bm{1}}

\DeclareMathAlphabet{\mathsfit}{\encodingdefault}{\sfdefault}{m}{sl}
\SetMathAlphabet{\mathsfit}{bold}{\encodingdefault}{\sfdefault}{bx}{n}

\newcommand{\Ls}{\mathcal{L}}

\newcommand{\norm}[1]{\ensuremath{\| #1 \|_2}}

\usepackage{microtype}
\usepackage{graphicx}
\usepackage{booktabs} 

\usepackage{subcaption}

\usepackage{amsmath}
\usepackage{amssymb}
\usepackage{mathtools}
\usepackage{amsthm}
\usepackage{epsfig}
\usepackage{graphicx}
\usepackage{amsmath}
\usepackage{amssymb}
\usepackage{enumitem}
\usepackage{xcolor}
\usepackage{multirow}
\usepackage{booktabs}
\usepackage{bbm}
\usepackage{caption}
\usepackage{subcaption}
\usepackage{siunitx}
\usepackage{tikz}
\usetikzlibrary{bayesnet}
\usetikzlibrary{arrows}
\usetikzlibrary{shapes.geometric, arrows}
\usepackage{color}
\usepackage{graphicx}
\usepackage{subcaption}
\usetikzlibrary{backgrounds}
\usepackage{array}
\usepackage{mathtools}
\usepackage{xcolor}
\usepackage{soul}

\theoremstyle{plain}

\theoremstyle{definition}

\theoremstyle{remark}

\begin{document}

\maketitle

\begin{abstract}
  Modeling spatial-temporal interactions among neighboring agents is at the heart of multi-agent problems such as motion forecasting and crowd navigation.
Despite notable progress, it remains unclear to which extent modern representations can capture the causal relationships behind agent interactions.
In this work, we take an in-depth look at the causal awareness of these representations, from computational formalism to real-world practice.
First, we revisit the notion of non-causal robustness studied in the recent CausalAgents benchmark \citep{roelofsCausalAgentsRobustnessBenchmark2022a}.
We show that existing representations are already partially resilient to perturbations of non-causal agents, and yet modeling indirect causal effects involving mediator agents remains challenging.
To address this challenge, we introduce a metric learning approach that regularizes latent representations with causal annotations.
Our controlled experiments show that this approach not only leads to higher degrees of causal awareness but also yields stronger out-of-distribution robustness.
To further operationalize it in practice, we propose a sim-to-real causal transfer method via cross-domain multi-task learning.
Experiments on trajectory prediction datasets show that our method can significantly boost generalization, even in the absence of real-world causal annotations, where we acquire higher prediction accuracy by only using 25\% of real-world data.
We hope our work provides a new perspective on the challenges and pathways toward causally-aware representations of multi-agent interactions. 
Code is available at \url{https://github.com/vita-epfl/CausalSim2Real}.
\end{abstract}

\section{Introduction}
\label{sec:intro}
Modeling multi-agent interactions with deep neural networks has made great strides in the past few years~\citep{alahiSocialLSTMHuman2016,vemulaSocialAttentionModeling2018,guptaSocialGANSocially2018a,kosarajuSocialBiGATMultimodalTrajectory2019,salzmannTrajectronDynamicallyFeasibleTrajectory2020a,mangalamGoalsWaypointsPaths2021,guDenseTNTWaymoOpen2021,xuPreTraMSelfsupervisedPretraining2022,chenUnsupervisedSamplingPromoting2023}.
Yet, existing representations still face tremendous challenges in handling changing environments: they often suffer from substantial accuracy drops under mild environmental changes~\citep{liuSocialNCEContrastive2021,bagiGenerativeCausalRepresentation2023} and require a large number of examples for adaptation to new contexts~\citep{moonFastUserAdaptation2022,kothariMotionStyleTransfer2022}.
These challenges are arguably rooted in the nature of the learning approach that seeks statistical correlations in the training data, regardless of their stability and reusability across distributions~\citep{castriCausalDiscoveryDynamic2022,bagiGenerativeCausalRepresentation2023}.
One promising approach is to develop {\it causally-aware representations}—latent features that capture the invariant causal dependencies in agent interactions.

\begin{figure}[t]
\vskip 0.1in
\begin{center}
\includegraphics[width=1.0\linewidth]{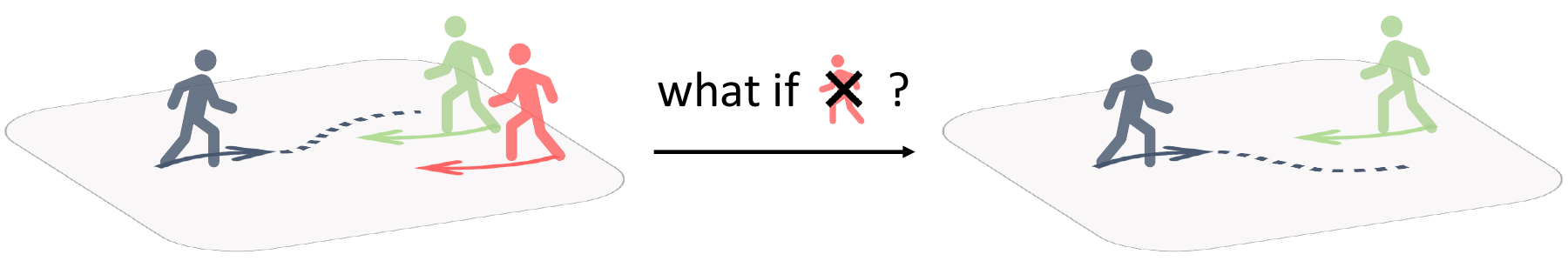}
\caption{\textbf{Illustration of causal relations in multi-agent interactions.} The behavior of the ego agent is causally influenced by some neighboring agents. 
Our research explores the depth of causal understanding within trajectory prediction models, introduces novel causal regularizers to refine this understanding, and details a sim-to-real approach that effectively transfers this causal knowledge to practical real-world scenarios.
}
\label{fig:pull}
\end{center}
\vskip -0.25in
\end{figure}

However, discovering causal knowledge from observational data is often exceptionally difficult~\citep{scholkopfCausalRepresentationLearning2021,scholkopfStatisticalCausalLearning2022}.
Most prior works resort to additional information, such as structural knowledge~\citep{chenHumanTrajectoryPrediction2021,makansiYouMostlyWalk2021} and domain labels~\citep{liuRobustAdaptiveMotion2022,huCausalbasedTimeSeries2022,bagiGenerativeCausalRepresentation2023}.
While these attempts have been shown effective in certain out-of-distribution scenarios, they still fall short of explicitly accounting for causal relationships between interactive agents.
More recently, CausalAgents~\citep{roelofsCausalAgentsRobustnessBenchmark2022a} made an effort to collect annotations of agent relations (causal or non-causal) in the Waymo dataset~\citep{ettingerLargeScaleInteractive2021}, providing a new benchmark focused on the robustness issue under non-causal agent perturbations.
Nevertheless, the reason behind the robustness issue remains unclear, as does the potential use of the collected annotations for representation learning.

The goal of this work is to provide an in-depth analysis of the challenges and opportunities of learning causally-aware representations in the multi-agent context.
To this end, we first take an in-depth look at the recent CausalAgents benchmark~\cite {roelofsCausalAgentsRobustnessBenchmark2022a}.
We find that its labeling procedure and evaluation protocol present subtle yet critical caveats, thereby resulting in a highly biased measure of robustness.
To mitigate these issues, we construct a diagnostic dataset through counterfactual simulations and revisit a collection of recent multi-agent forecasting models.
Interestingly, we find that most recent models are already partially resilient to perturbations of non-causal agents but still struggle to capture indirect causal effects that involve mediator agents.

To further enhance the causal robustness of the learned representations, we propose a regularization approach that seeks to preserve the causal effect of each individual neighbor in an embedding space.
Specifically, we devise two variants that exploit annotations with different levels of granularity:
(i) a contrastive-based regularizer using binary annotations of causal/non-causal agents; (ii) a ranking-based regularizer using continuous annotations of causal effects.
Through controlled experiments, we show that both regularizers can enhance causal awareness to notable degrees.
More crucially, we find that finer-grained annotations are particularly important for generalization out of the training distribution, such as higher agent density or unseen context.

Finally, we introduce a sim-to-real causal transfer framework, aiming at extending the strengths of causal regularization from simulation to real-world contexts.
We achieve this through cross-domain multi-task learning, \ie, jointly train the representation on the causal task in simulation and the forecasting task in the real world.
Through experiments on the ETH-UCY dataset~\citep{lernerCrowdsExample2007,pellegriniImprovingDataAssociation2010} paired with an ORCA simulator~\citep{vandenbergReciprocalVelocityObstacles2008}, we find that the causal transfer framework enables stronger generalization in challenging settings such as low-data regimes. 
Despite a large gap between simulator and real-world data, our approach outperforms the AutoBots baseline with only 25\% of the real data when combined with our causal regularizers. Furthermore, with just 50\% of the real data, it surpasses all other models in prediction accuracy. This is achieved even in the absence of real-world causal annotations, underscoring the robustness of our framework in limited-data scenarios. We also test the causal transfer approach on the NBA dataset \cite{xu2022socialvae,yue2014learning}, demonstrating its effectiveness in different real-world settings. As one of the first steps towards causal modeling in multi-agent systems, we hope our work sheds new light on the challenges and opportunities of learning causally-aware representations in practice.

\section{Related Work}
\label{sec:related}
The social causality studied in this work lies at the intersection of three areas: multi-agent interactions, robust representations, and causal learning.
In this section, we provide a brief overview of the existing literature in each area and then discuss their relevance to our work.

\noindent\textbf{Multi-Agent Interactions.}
The study of multi-agent interactions has a long history. Early efforts were focused on hand-crafted rules, such as social forces~\citep{helbingSocialForceModel1998, mehranAbnormalCrowdBehavior2009} and reciprocal collision avoidance~\citep{vandenbergReciprocalVelocityObstacles2008, alahiSociallyAwareLargeScaleCrowd2014}.
Despite remarkable results in sparse scenarios~\citep{luberPeopleTrackingHuman2010,zanlungoSocialForceModel2011, ferrerRobotCompanionSocialforce2013}, these models often lack social awareness in more densely populated and complex environments~\citep{rudenkoHumanMotionTrajectory2020}.
As an alternative, recent years have witnessed a paradigm shift toward learning-based approaches, particularly the use of carefully designed neural networks to learn representations of multi-agent interactions~\citep{kothariHumanTrajectoryForecasting2021}.
Examples include pooling operators~\citep{alahiSocialLSTMHuman2016,guptaSocialGANSocially2018a, deoConvolutionalSocialPooling2018}, attention mechanisms~\citep{vemulaSocialAttentionModeling2018,sadeghianSoPhieAttentiveGAN2019,huangSTGATModelingSpatialTemporal2019,gao2024multi}, spatio-temporal graphs~\citep{kosarajuSocialBiGATMultimodalTrajectory2019,salzmannTrajectronDynamicallyFeasibleTrajectory2020a,liEvolveGraphMultiAgentTrajectory2020}, among others~\citep{rhinehartPRECOGPREdictionConditioned2019,chaiMultiPathMultipleProbabilistic2020,choiDROGONTrajectoryPrediction2021}.
Nevertheless, the robustness of these models remains a grand challenge~\citep{saadatnejadAreSociallyawareTrajectory2022,roelofsCausalAgentsRobustnessBenchmark2022a,caoRobustTrajectoryPrediction2023}.
Our work presents a solution to enhance robustness by effectively exploiting causal annotations of varying granularity.

\noindent\textbf{Robust Representations.} The robustness of machine learning models, especially under distribution shifts, has been a long-standing concern for safety-critical applications~\citep{rechtImageNetClassifiersGeneralize2019}. Existing efforts have explored two main avenues to address this challenge. One line of work seeks to identify features that are invariant across distributions. Unfortunately, this approach often relies on strong assumptions about the underlying shifts~\citep{hendrycks*AugMixSimpleData2020,liuSocialNCEContrastive2021} or on access to multiple training domains~\citep{arjovskyInvariantRiskMinimization2020,kruegerOutofDistributionGeneralizationRisk2021}, which may not be practical in real-world settings. Another approach aims to develop models that can efficiently adapt to new distributions by updating only a small number of weight parameters, such as sub-modules~\citep{kothariMotionStyleTransfer2022}, certain layers~\citep{leeSurgicalFineTuningImproves2023}, or a small subset of neurons~\citep{chenRevisitingParameterEfficientTuning2022}. More recently, there has been a growing interest in exploring the potential of causal learning to address the robustness challenges~\citep{vansteenkisteAreDisentangledRepresentations2019,dittadiTransferDisentangledRepresentations2020,monteroRoleDisentanglementGeneralisation2022,dittadiGeneralizationRobustnessImplications2022a}.
To the best of our knowledge, our work makes the first attempt in the multi-agent context, showcasing the benefits of incorporating causal relationships for stronger out-of-distribution generalization in crowded scenarios.

\noindent\textbf{Causal Learning.} Empowering machine learning with causal reasoning has gained a growing interest in recent years~\citep{scholkopfCausalityMachineLearning2019,scholkopfStatisticalCausalLearning2022}.
One line of work seeks to discover high-level causal variables from low-level observations, \eg, disentangled~\citep{chenInfoGANInterpretableRepresentation2016,higginsBetaVAELearningBasic2017,locatelloChallengingCommonAssumptions2019} or structured latent representations~\citep{locatelloObjectCentricLearningSlot2020a,scholkopfCausalRepresentationLearning2021}. Unfortunately, existing methods remain largely limited to simple and static datasets~\citep{liuCausalTripletOpen2023}.
Another thread of work attempts to draw causal insights into dynamic decision-making~\citep{huangAdaRLWhatWhere2022}. In particular, recent works have proposed a couple of methods to incorporate causal invariance and structure into the design and training of forecasting models in the multi-agent context~\citep{chenHumanTrajectoryPrediction2021,makansiYouMostlyWalk2021,liuRobustAdaptiveMotion2022,huCausalbasedTimeSeries2022,castriCausalDiscoveryDynamic2022,bagiGenerativeCausalRepresentation2023}. However, these efforts have mainly focused on using causal implications to enhance robustness rather than explicitly examining causal relations between interactive agents.
Closely related to ours, another recent study introduces a motion forecasting benchmark with annotations of causal relationships~\citep{roelofsCausalAgentsRobustnessBenchmark2022a}.
Our work addresses this benchmark's limitations and introduces a sim-to-real approach to enhance the causal awareness of the learned representations.

\section{Preliminaries}
\label{sec:annot}
In this section, we seek to formalize the robustness challenge in multi-agent representation learning through the lens of social causality. We will first revisit the design of the CausalAgents~\citep{roelofsCausalAgentsRobustnessBenchmark2022a} benchmark, drawing attention to its caveats in labeling and evaluation. Subsequently, we will introduce a diagnostic dataset, aiming to facilitate more rigorous development and evaluation of causally-aware representations through counterfactual simulations. 

\subsection{Social Interaction and Social Causality}

\begin{figure*}[t]
\centering
\begin{minipage}{0.48\textwidth}
  \centering
  \includegraphics[width=0.72\linewidth]{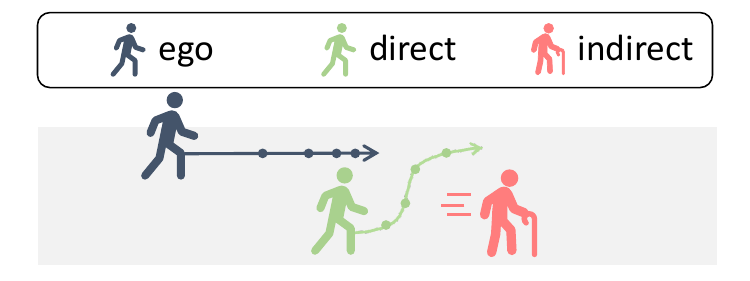}
  \caption{\textbf{Illustration of indirect causal effects.} The elderly pedestrian \textit{indirectly} affects the trajectory of the ego agent by influencing the path of the green pedestrian. In the elderly pedestrian's absence, the green pedestrian would proceed straight, allowing the ego agent to continue at its current speed without needing to slow down.}
  \label{fig:caveat_anno}
\end{minipage}
\hfill
\begin{minipage}{0.48\textwidth}
  \centering
  \includegraphics[width=0.57\linewidth]{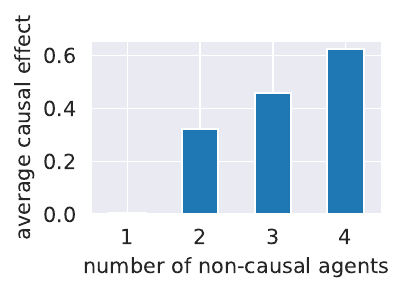}
  \caption{\textbf{Joint effect of removing all non-causal agents in the scene, simulated in ORCA.} The ego agent often changes its behavior when more than 2 non-causal neighbors are removed.}
  \label{fig:caveat_eval}
\end{minipage}
\vskip -0.1in
\end{figure*}
\textbf{Social Interaction.} Consider a motion forecasting problem where an ego agent is surrounded by a set of neighboring agents in a scene. 
Let $s_t^i = (x_t^i, y_t^i)$ denote the state of agent $i$ at time $t$ and $s_t = \{s_t^0, s_t^1, \cdots, s_t^K\}$ denote the joint state of all the agents in the scene.
Without loss of generality, we index the ego agent as $0$ and the rest of neighboring agents as $\mathcal{A} = \{1, 2, \cdots, K\}$.
Given a sequence of history observations $\bm x=(s_1, \cdots, s_t)$, the task is to predict the future trajectory of the ego agent $\bm y = (s_{t+1}^0, \cdots, s_{T}^0)$ until time $T$.
Modern forecasting models are largely composed of encoder-decoder neural networks, where the encoder $f(\cdot)$ first extracts a compact representation $\bm z$ of the input with respect to the ego agent and the decoder $g(\cdot)$ subsequently rolls out a sequence of its future trajectory $\bm{\hat{y}}$:
\begin{equation}
\begin{split}\label{eq:encoder_decoder}
\bm z &= f(\bm x) = f(s_{1:t}),  \\
\hat {\bm y} &= \hat {s}_{t+1:T}^0 = g(\bm z).
\end{split}
\end{equation}

\noindent\textbf{Social Causality.}
Despite remarkable progress on accuracy measures~\citep{alahiSocialLSTMHuman2016,salzmannTrajectronDynamicallyFeasibleTrajectory2020a}, recent neural representations of social interactions still suffer from a significant robustness concern. For instance, recent works have shown that trajectories predicted by existing models often output colliding trajectories~\citep{liuSocialNCEContrastive2021}, vulnerable to adversarial perturbations~\citep{saadatnejadAreSociallyawareTrajectory2022,caoAdvDORealisticAdversarial2022} and deteriorate under distribution shifts of spurious feature such as agent density~\citep{liuRobustAdaptiveMotion2022}. One particular notion of robustness that has recently gained much attention is related to the causal relationships between interactive agents~\citep{roelofsCausalAgentsRobustnessBenchmark2022a}.
Ideally, neural representations of social interactions should capture the influence of neighboring agents, namely how the future trajectory of the ego agent will vary between a pair of scenes: an original scene $\bm x_\varnothing$ and a perturbed scene $\bm x_R$ where some neighboring agents $\mathcal{R}$ are removed, \ie, only the ego agent and $\mathcal{A} \setminus \mathcal{R}$ remain. We define the causal effect of the removed agents $\mathcal{R}$ as,
\begin{equation}\label{eq:causality}
    \mathcal{E}_\mathcal{R} = \|{\bm {y}_\varnothing - \bm {y}_\mathcal{R}}\|_2,
\end{equation}
where $\bm {y}_\varnothing \equiv \bm {y}$ is the future trajectory in the original scene, $\bm {y}_\mathcal{R}$ is the future trajectory in the perturbed scene after removing agents in $\mathcal{R}$ from scene, and $\|{\cdot}\|_2$ is the average point-wise Euclidean distance between two trajectories.


\subsection{Caveats of CausalAgents Benchmark} \label{subsec:caveat}

While the mathematical definition of the causal effect is straightforward in the multi-agent context, measuring \cref{eq:causality} in practice can be highly difficult. In general, it is impossible to observe a subset of agents replaying their behaviors in the same environment twice in the real world -- an issue known as the impossibility of counterfactuals~\citep{petersElementsCausalInference2017}. To mitigate the data collection difficulty, a recent benchmark CausalAgents~\citep{roelofsCausalAgentsRobustnessBenchmark2022a} proposes another simplified labeling strategy: instead of collecting paired counterfactual scenes, it queries human labelers to divide neighboring agents into two categories: causal agents that directly influence the driving behavior of the ego agent from camera viewpoints; and non-causal agents that do not. This labeling approach is accompanied by an evaluation protocol through agent removal, assuming that robust forecasting models should be insensitive to scene-level perturbations that remove non-causal agents. Formally, the CausalAgents benchmark evaluates the robustness of a learned representation through the following measure,
\begin{equation}\label{eq:waymo}
\begin{split}
    \Delta =
    \|{\bm {\hat y}_\mathcal{R} - \bm {y}_\mathcal{R}}\|_2 - 
    \|{\bm {\hat y}_\varnothing - \bm {y}_\varnothing}\|_2,
\end{split}
\end{equation}
where $\mathcal{R}$ includes all or some non-causal agents in the scene.


\noindent\textbf{Caveats.} In spite of the precious efforts in gathering causal annotations, the CausalAgents benchmark~\citep{roelofsCausalAgentsRobustnessBenchmark2022a} is subject to two limitations that require attention:
\begin{enumerate}[nosep]
    \item {\em Annotation}: The human labeling process overlooks indirect causal effects. As illustrated in~\cref{fig:caveat_anno}, a neighboring agent may have a considerable causal influence on the behavior of the ego agent via mediators. 
    The complexity of these causal chains poses a significant challenge for human annotations. Yet, these indirect effects are common in scenarios with high agent density, presenting a significant modeling challenge, which we will demonstrate in \cref{sec:experiment}.
    \item {\em Evaluation}: 
    The evaluation protocol of the CausalAgents benchmark may inaccurately portray robustness due to oversimplification in handling agent interactions, as illustrated in ~\cref{fig:caveat_eval}. This figure shows that the impact of non-causal agents grows when they are removed collectively rather than individually. While agents are marked individually as non-causal, the robustness test removes them in groups, which can exaggerate the model's perceived robustness. Simultaneous removal of multiple non-causal agents might expose causal effects that are not detected when these agents are considered separately. For a detailed scenario illustrating this effect, please refer to \cref{appendix:caveat_eval}.
\end{enumerate}

    
\subsection{Diagnostic Dataset through Counterfactuals} \label{sec:dataset}
To address the identified issues, we have developed a new diagnostic dataset using the ORCA~\citep{vandenbergReciprocalVelocityObstacles2008} simulation system, which enables the creation of counterfactual scenarios by removing sets of agents. ORCA models social interactions like collision avoidance and leader-follower behavior by taking initial positions, speeds, and final goals of agents. We constructed our dataset by initializing the positions and goals of $8 \leq K+1 \leq 18$ agents within a structured layout typical of an open area. Additionally, we generated two out-of-distribution (OOD) datasets, named Density OOD and Context OOD, to test model robustness. The Density OOD dataset varies the density of agents and adds more non-causal ones, while the Context OOD dataset alters the initial layout to resemble a narrow street with pedestrians moving from one end to the other. Both the datasets and the code used for their generation are available to support further research. For more detailed information on the data collected, see \cref{appendix:implementation}.

\noindent \textbf{Counterfactual Pairs.} Leveraging the capability to simulate counterfactual scenarios, we annotate the ground-truth causal effects by comparing ego trajectories in paired scenes before and after agent removals, as defined in~\cref{eq:causality}. Given the exponential growth of simulation scenarios with the number of agents, we adopt the approach of Causal Agents~\citep{roelofsCausalAgentsRobustnessBenchmark2022a} and concentrate on evaluating causal effects at the level of individual agents, denoted as \ie $|\mathcal{R}| = 1$. For simplicity, moving forward, we will use $\mathcal{E}_i$ to refer to $\mathcal{E}_{\{i\}}$.

\noindent \textbf{Fine-grained Category.} In addition to real-valued causal effects, we also seek to annotate the category of each agent.
From simulations, we extract the per-step causal relation between a neighboring agent $i$ and the ego agent. If agent $i$ is visible to the ego at time step $t$, it directly influences the ego, indicated by $\mathbbm{1}^i_t=1$; otherwise $\mathbbm{1}^i_t=0$.
We then convert the causal effect over the whole sequence $\mathcal{E}_i$ into three agent categories:
\begin{itemize}[nosep]
    \item {\em Non-causal agent}: little influence on the behavior of the ego agent, \ie, $\mathcal{E} < \epsilon \approx 0$.
    \item {\em Direct causal agent}: significant influence on the behavior of the ego agent $\mathcal{E} > \eta \gg 0$; moreover, the influence is direct for at least one time step $\prod_{\tau={1:T}} (1-\mathbbm{1}^i_\tau) = 0$.
    \item {\em Indirect causal agent}: significant influence on the behavior of the ego agent $\mathcal{E} > \eta \gg 0$; however, the influence is never direct over the entire sequence $\prod_{\tau={1:T}} (1-\mathbbm{1}^i_\tau) = 1$.
\end{itemize}
The collected diagnostic dataset with different levels of causal annotations allows us to rigorously probe the causal robustness of existing representations and to develop more causally-aware representations, which we will describe in the next sections.

\section{Method}
\label{sec:method}
\begin{figure*}[ht]
\begin{center}
\centerline{\includegraphics[width=0.8\textwidth]{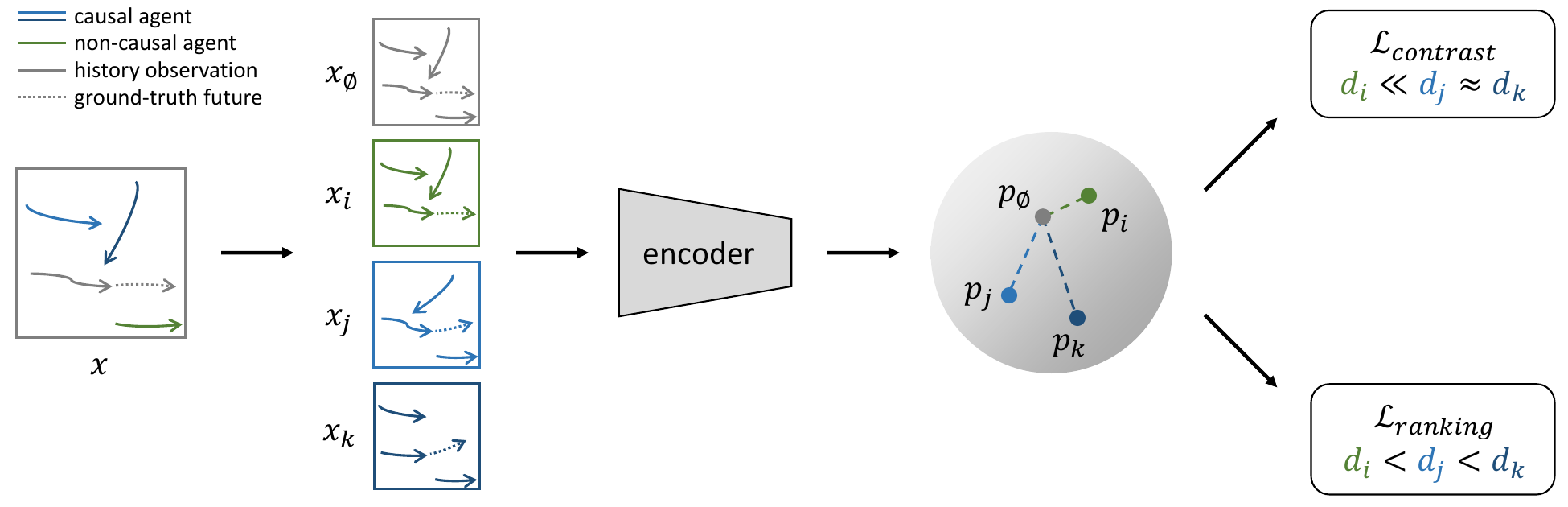}}
\caption{\textbf{Overview of our method.} We seek to build an encoder that captures the causal effect of each agent by regularizing the distance of paired embeddings between the factual and counterfactual scenes.
  We formulate this objective into a contrastive task given binary annotations or a ranking task given real-value annotations.}
\label{fig:method}
\end{center}
\vskip -0.15in
\end{figure*}

In this section, we will introduce a simple yet effective approach to promote causal awareness of the learned representations.
We will first describe a general regularization method, which encapsulates two specific instances exploiting causal annotations of different granularity. 
We will then introduce a sim-to-real transfer framework that extends the regularization method to practical settings, even in the absence of real-world annotations.

\subsection{Causally-Aware Regularization}

Recall that the causal effect $\mathcal{E}_i$ of an agent $i$ is tied to the {\em difference} of the potential outcomes between the factual scene and the counterfactual one where the agent $i$ is removed.
In this light, a causally-aware representation should also capture such relations, \ie, feature vectors of paired scenes are separated by a certain distance $d_i$ depending on $\mathcal{E}_i$.

We start by defining $\bm p_\varnothing$  as the low-dimensional feature vector of the factual scene where no agents are removed. This serves as our baseline scenario against which all counterfactual instances are compared. $\bm p_\varnothing$ is extracted from the latent representation $\bm z_\varnothing=f(\bm x_\varnothing)$ projected through a non-linear head $h$. Similarly, $\bm p_i = h (\bm z_i) = h (f(\bm x_i))$ the feature vector for the counterfactual scene $\bm x_i$ created by removing agent $i$.
We then measure the distance between $\bm x_\varnothing$ and $\bm x_i$ as
\begin{equation}
    d_i = 1 - \textrm{sim} (\bm p_\varnothing, \bm p_i) = 1 - \frac{\bm p_\varnothing^\top \bm p_i} 
    {\lVert \bm p_\varnothing \rVert \lVert \bm p_i \rVert},
\end{equation}
where $\textrm{sim}(\bm p_\varnothing, \bm p_i)$ denotes the cosine similarity between the feature vectors of the factual and counterfactual scenes, effectively measuring the impact of agent $i$'s removal on the embedding space.
The desired representation is thus expected to preserve the following property,
\begin{equation} \label{eq:compare}
    \mathcal{E}_i < \mathcal{E}_j \implies d_i < d_j, \quad \forall i, j \in \mathcal{A},
\end{equation}
where $i$ and $j$ are the indices of two agents in the set of neighboring agents $\mathcal{A}$.
We next describe two specific tasks that seek to enforce this property~\cref{eq:compare}, while taking into account the concrete forms of causal annotations with different granularity, as illustrated in \cref{fig:method}.

\noindent\textbf{Causal Contrastive Regularization.}
As discussed in \cref{sec:dataset}, one simple form of causal annotations is binary labels, indicating whether or not a neighboring agent causally influences the behavior of the ego agent.
Intuitively, the embedding of a counterfactual scenario where a non-causal agent is removed should closely resemble the embedding of the factual scenario, reflecting minimal change from the ego agent's perspective. In contrast, removing a causal agent should significantly alter the representation, indicating a higher impact.
We formulate this intuition into a causal contrastive learning objective,
\begin{equation}
    \Ls_{\mathrm{contrast}} = - \log \frac{\exp (d^+ / \tau)}{\exp (d^+ / \tau) + \sum_k \exp (d_k / \tau) \mathbbm{1}_{\mathcal{E}_k > \eta} },
\end{equation}
where the positive (distant) example is sampled from counterfactuals with respect to causal agents, the negative (nearby) examples are sampled from counterfactual scenarios with respect to non-causal agents, and $\tau$ is a temperature hyperparameter controlling the difficulty of the contrastive task.

\noindent\textbf{Causal Ranking Regularization.}
One downside of causal contrastive learning described above is that it inherently ignores the detailed effect of causal agents.
It tends to push the embeddings of counterfactuals equally far apart across all causal agents, regardless of the variation of causal effects, which violates the desired property stated in \cref{eq:compare}.
To address this limitation, we further consider another variant of causal regularization using real-valued causal annotations to provide more dedicated supervision on the relative distance in the embedding space.
Concretely, we first sort all agents in a scene based on their causal effect and then sample two agents with different causal effects for comparison.
This allows us to formulate a ranking problem in a pairwise manner through a margin ranking loss,
\begin{equation} \label{eq:loss}
    \Ls_{\mathrm{ranking}} = \max(0, d_i - d_j + m),
\end{equation}
where $d_i$ and $d_j$ are the embedding distances with respect to two agents of different causal effects $\mathcal{E}_i < \mathcal{E}_j$, and $m$ is a small margin hyperparameter controlling the difficulty of the ranking task.

\begin{table*}[t]
\begin{center}
\begin{small}
\begin{sc}
    \begin{tabular}{l c c c c c}
    \toprule
            & ADE {$\downarrow$} & FDE {$\downarrow$} & ACE-NC {$\downarrow$} & ACE-DC {$\downarrow$} & ACE-IC {$\downarrow$}\\
    \midrule
    D-LSTM~\citep{kothariHumanTrajectoryForecasting2021} & 0.329 & 0.677 & 0.027  & 0.532 & 0.614 \\
    S-LSTM~\citep{alahiSocialLSTMHuman2016} & 0.314  & 0.627& 0.031 & 0.463 & 0.523  \\
    Trajectron++~\cite{salzmannTrajectronDynamicallyFeasibleTrajectory2020a} & 0.312 & 0.630 & 0.024 & 0.479 & 0.568 \\
    STGCNN~\citep{mohamedSocialSTGCNNSocialSpatioTemporal2020} & 0.307 & 0.564 & 0.049 & 0.330 & 0.354\\
    Multi-Transmotion~\citep{gao2024multi} & 0.237 & 0.466 & 0.031 & 0.723 & 0.724\\
    AutoBots~\citep{girgisLatentVariableSequential2021} & 0.255 & 0.497 & 0.045 & 0.595 & 0.616 \\
    \bottomrule
    \end{tabular}
\end{sc}
\end{small}
\caption{\textbf{Performance of modern representations on the created diagnostic dataset.}
The prediction errors and causal errors are generally considerable across all evaluated models. However, the errors associated with ACE-NC are rather marginal, suggesting that recent models are already partially robust to non-causal perturbations of individual agents.}
\label{tab:baselines}
\end{center}
\vskip -0.1in
\end{table*}

\subsection{Sim-to-real Causal Transfer}

The causal regularization method described above relies upon the premise that annotations of causal effects are readily available. However, as elaborated in~\cref{sec:dataset}, procuring such causal annotations in real-world scenarios can be difficult.
To bridge this gap, we extend our causal regularization approach to a sim-to-real transfer learning framework.
Our key idea is that, despite the gap between simulation environments and the real world, certain causal mechanisms (\eg, collision avoidance and group coordination) tend to remain stable.
As such, in the absence of real-world causal annotations, we can use the annotations gathered from simulations as a viable alternative. 
This leads to cross-domain training on two tasks jointly:
the main prediction task $\Ls_{pred}^{real}$ in real-world scenarios which usually is a reconstruction loss, and our auxiliary causal task $\Ls_{causal}^{syn}$ in the simulation counterpart,
\begin{equation}
    \Ls = \Ls_{\mathrm{pred}}^{\mathrm{real}} + \alpha \Ls_{\mathrm{causal}}^{\mathrm{syn}},
\end{equation}
where $\alpha$ is a hyperparameter controlling the emphasis on the causal regularizer.
Despite its simplicity, we will show in~\cref{subsec:s2r} that the sim-to-real causal transfer framework effectively translates the causal knowledge absorbed from simulations to real-world datasets, even without any real-world causal annotations.



\section{Experiments}
\label{sec:experiment}

In this section, we will present a set of experiments to answer the following four questions:
\begin{enumerate}[nosep]
    \item How well do recent representations capture the causal relations between interactive agents?
    \item Is our proposed method effective for addressing the limitations of recent representations?
    \item Do finer-grained annotations provide any benefits for learning causally-aware representations?
    \item Finally, does greater causal awareness offer any practical advantages in challenging scenarios?
\end{enumerate}

Throughout our experiments, the multi-agent forecasting task is defined as predicting the future trajectory of the ego agent for 12 time steps, given the history observations of all agents in a scene in the past 8 time steps. Given that our simulator operates deterministically without random variations in agent trajectories, we train our forecasting models to output a single future trajectory for the ego agent, rather than generating multiple potential trajectories in multimodal settings.
We evaluate forecasting models on three metrics:
\begin{itemize}[nosep]
    \item \textit{Average Displacement Error} (ADE): the average Euclidean distance between the predicted output and the ground truth trajectories -- a  widely used metric to measure the prediction accuracy of a forecasting model.
    \item \textit{Final Displacement Error} (FDE): the Euclidean distance between the predicted output and the ground truth at the last time step -- another common metric measuring prediction accuracy.
    \item \textit{Average Causal Error} (ACE): the average difference between the estimated causal effect and the ground truth causal effect -- a metric inspired by~\citet{roelofsCausalAgentsRobustnessBenchmark2022a} to measure the causal awareness of a learned representation,
    \begin{equation}
        \mathrm{ACE} \coloneqq \frac{1}{K} \sum_{i=1}^K \lvert \mathcal{\hat E}_{i} - \mathcal{E}_{i} \rvert = \frac{1}{K} \sum_{i=1}^K \lvert \norm{\bm {\hat y}_\varnothing - \bm {\hat y}_{i}} - \mathcal{E}_{i} \rvert.
    \end{equation}
\end{itemize}
Alongside the aggregated ACE, we also measure causal awareness on each category separately, \ie, ACE-NC, ACE-DC, ACE-IC for non-causal, direct causal, and indirect causal agents, respectively.
Note that ACE can only be evaluated in the presence of causal annotations $\mathcal{E}_i$, \eg, the diagnostic dataset through controlled simulations (\cref{sec:dataset}).

\begin{figure}[t]
    \centering
    \begin{subfigure}[b]{0.485\linewidth}
        \centering
        \includegraphics[width=1.0\linewidth]{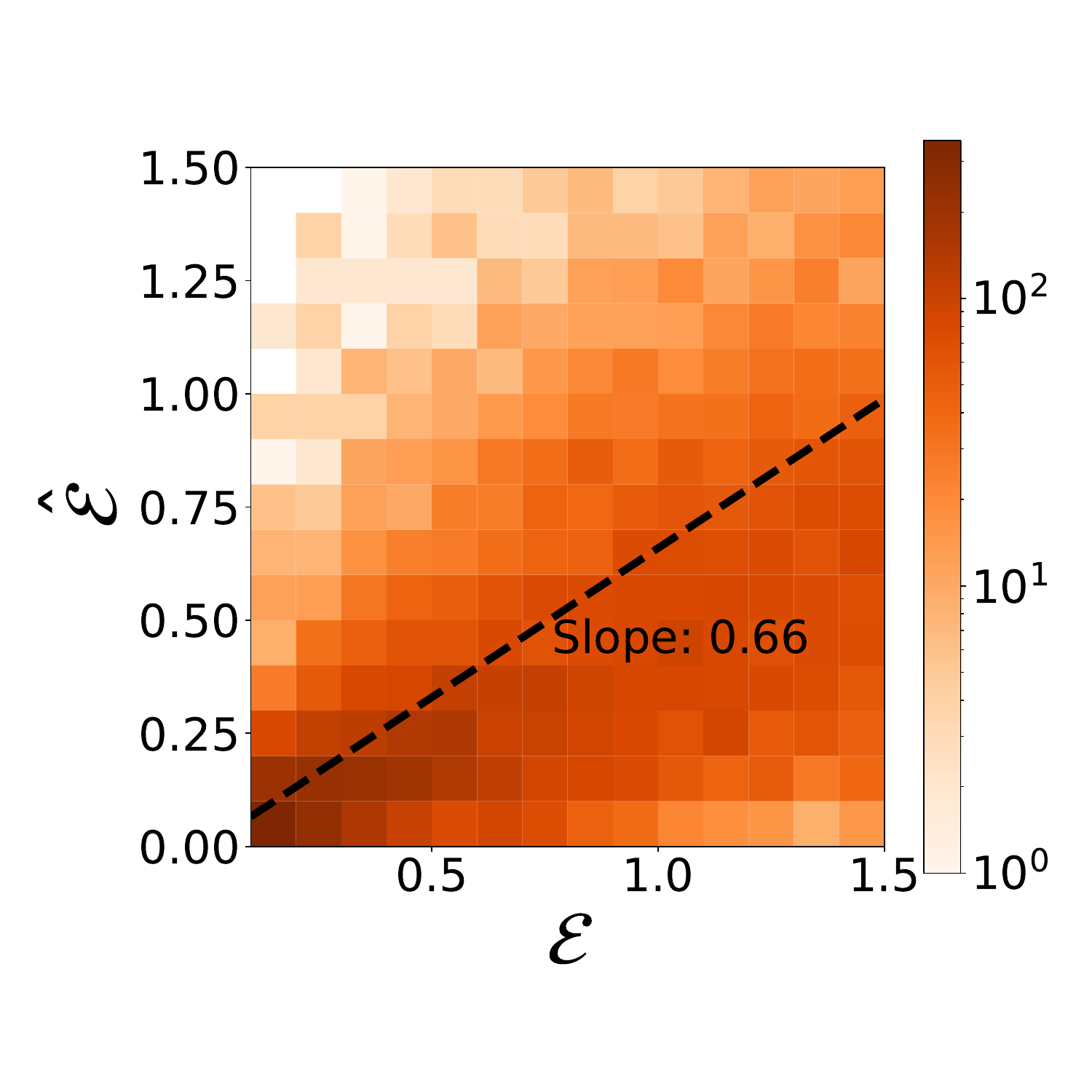}
        \vspace{-15pt}
        \caption{Direct Causal Agents}
        \label{fig:sensitivity_indirectcausal}
    \end{subfigure}
    ~
    \begin{subfigure}[b]{0.485\linewidth}
        \centering
        \includegraphics[width=1.0\linewidth]{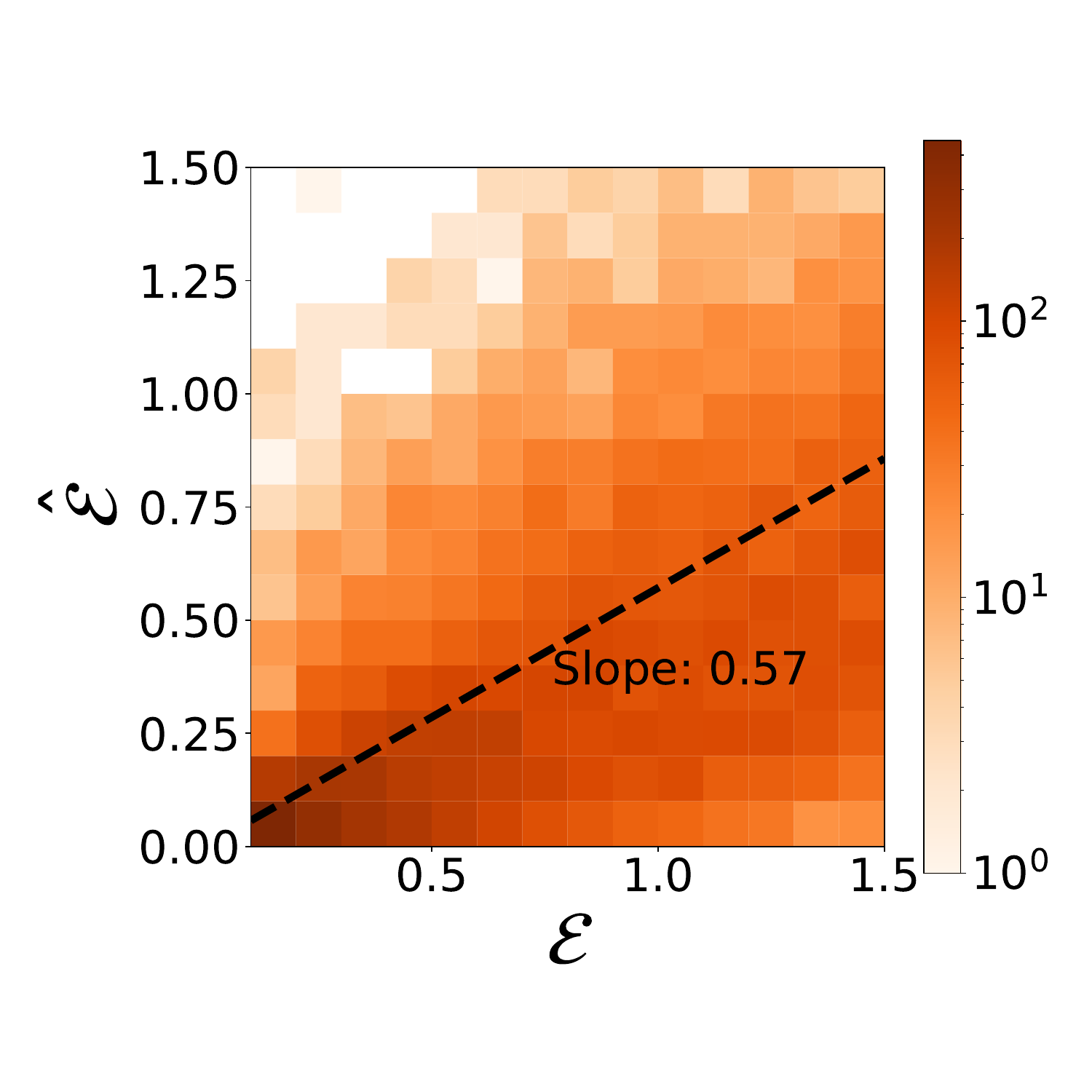}
        \vspace{-15pt}
        \caption{Indirect Causal Agents}
        \label{fig:sensitivity_directcausal}
    \end{subfigure}
    \caption{\textbf{Comparison between estimated causal effects $\mathcal{\hat E}$ and the ground truth $\mathcal{E}$.} We uniformly sample the ground-truth causal effect, collect the estimate from AutoBots~\citep{girgisLatentVariableSequential2021}, and regress a slope between the estimated value and the ground truth. 
    The positioning of most data points below the ideal diagonal line suggests a systematic underestimation of causal effects.
    }
    \label{fig:causaleffect}
\end{figure}

\begin{figure*}[t]
    \vskip 0.1in
    \centering
    \begin{subfigure}[b]{0.55\linewidth}
        \centering
        \includegraphics[height=130pt]{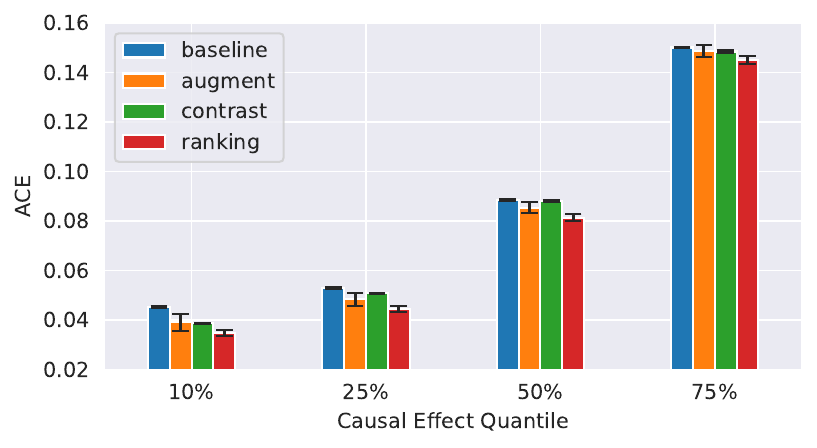}
    \end{subfigure}
    \quad
    \begin{subfigure}[b]{0.4\linewidth}
        \includegraphics[height=130pt]{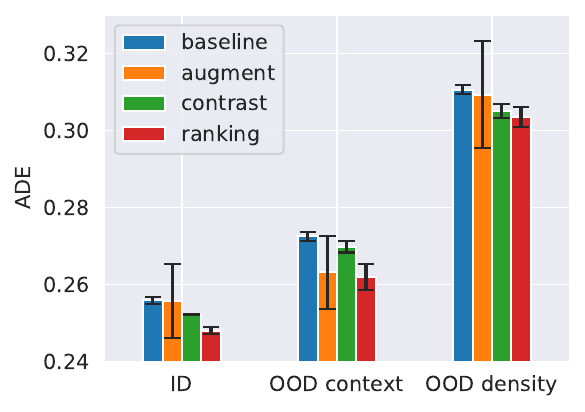}
    \end{subfigure}
  \caption{\textbf{Quantitaive results of our causal regularization method in comparison to the Autobots baseline~\citep{girgisLatentVariableSequential2021} and the non-causal data augmentation~\citep{roelofsCausalAgentsRobustnessBenchmark2022a}.} Models trained by our ranking-based method yield enhanced performance both in terms of both causal effect estimation (left) and prediction accuracy (right). Results are averaged over five random seeds.
  }
  \label{fig:quantitative} \label{fig:quantitative_ace}
  \label{fig:quantitative_ood}
  \vskip 0.0in
\end{figure*}

\begin{figure*}[t]
    \centering
    \begin{subfigure}[b]{0.32\linewidth}
        \centering
        \includegraphics[width=1.0\linewidth]{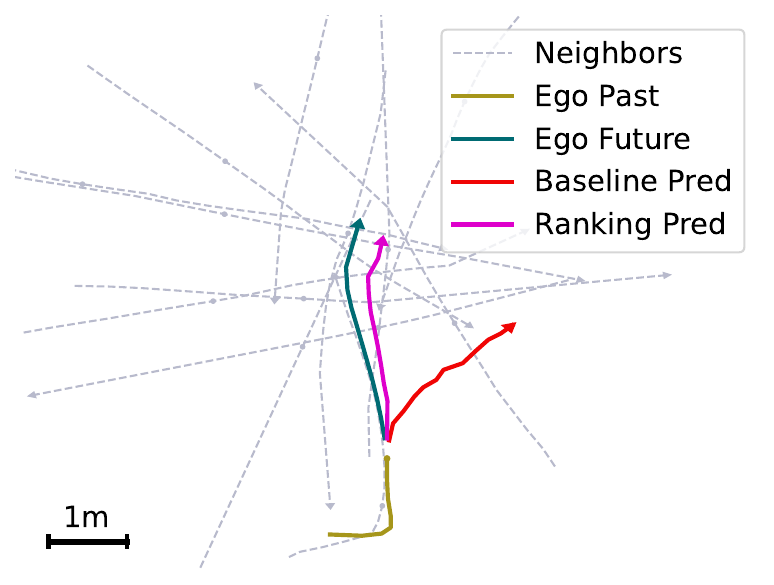}
        \caption{ID}
        \label{fig:visu_iid}
    \end{subfigure}
    ~
    \begin{subfigure}[b]{0.32\linewidth}
        \centering
        \includegraphics[width=1.0\linewidth]{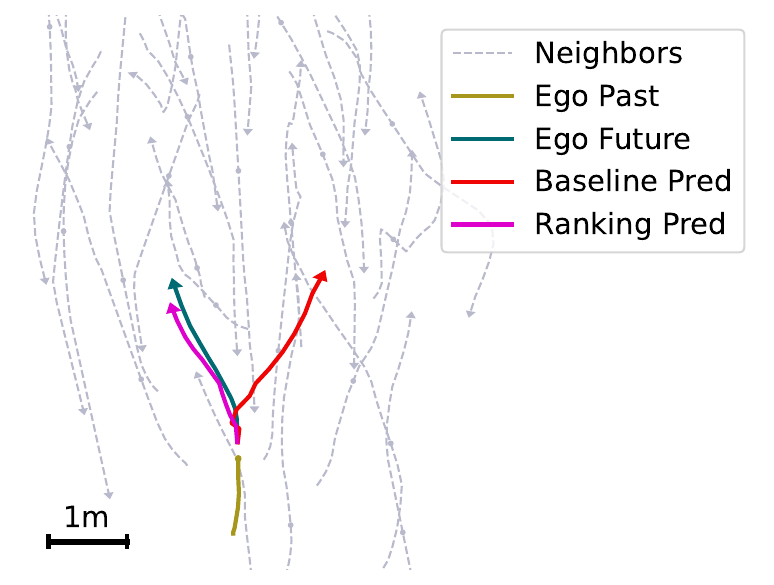}
        \caption{OOD-Context}
        \label{fig:visu_context}
    \end{subfigure}
    ~
    \begin{subfigure}[b]{0.32\linewidth}
        \centering
        \includegraphics[width=1.0\linewidth]{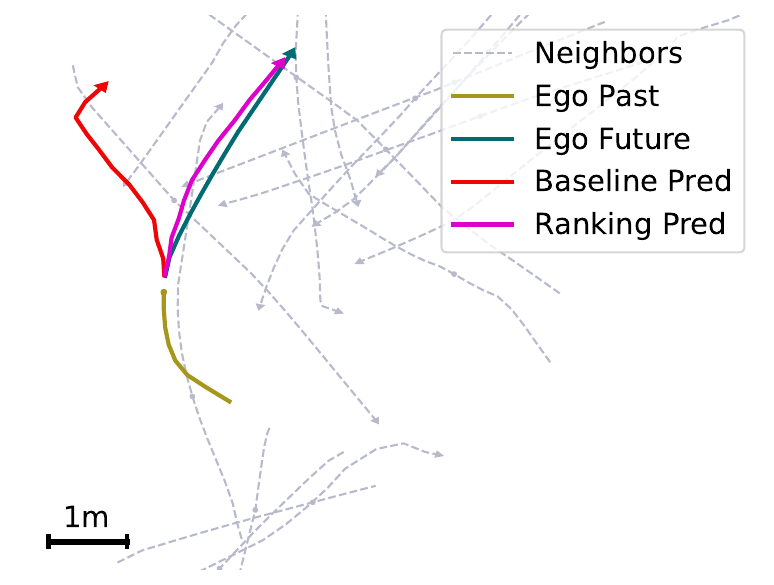}
        \caption{OOD-Density}
        \label{fig:visu_density}
    \end{subfigure}
    \caption{\textbf{Qualitative results of our method on in-distribution (ID) and out-of-distribution (OOD) test sets.} Models regularized by our ranking-based method show a more robust understanding of agent interactions in high-density and novel scene contexts.}
    \vskip -0.1in
    \label{fig:qualitative}
\end{figure*}

\subsection{Robustness of Recent Representations}
\label{subsec:recentrepresentation}

{\bf Setup.}
We start our experiments with the evaluation of a collection of recent models on the diagnostic dataset described in~\cref{sec:annot}.
We explicitly examine the errors of causal effect estimation for each agent category.
We summarize the results in~\cref{tab:baselines} and visualize the detailed estimation in~\cref{fig:causaleffect}.

\noindent{\bf Takeaway 1: recent representations are already partially robust to non-causal agent removal.}
As shown in~\cref{tab:baselines}, the values of ACE-NC are generally quite minimal compared to the results on other metrics.
Across all evaluated models, the errors made upon non-causal agents are 10x smaller than that for causal agents (ACE-DC/IC).
In line with our analysis in~\cref{sec:annot}, this result provides a counterargument to the robustness issue benchmarked in ~\citet{roelofsCausalAgentsRobustnessBenchmark2022a}, suggesting the importance of studying the robustness under causal perturbations as opposed to non-causal ones.

\noindent{\bf Takeaway 2: recent representations underestimate causal effects, particularly indirect ones.}
As shown in~\cref{fig:causaleffect}, the estimate of causal effect often deviates from the ground truth value.
In particular, the learned representation tends to severely underestimate the influence of indirect causal agents.
This result underscores the limitation of existing interaction representations in reasoning about a chain of causal relations transmitting across multiple agents.

\subsection{Effectiveness of Causal Regularization}
\label{subsec:synthetic}

We next evaluate the efficacy of our causal regularization method on Autobots, the strongest baselines in~\cref{tab:baselines}.
As a baseline comparison, we incorporate the data augmentation strategy from~\citep{roelofsCausalAgentsRobustnessBenchmark2022a}, which augments training scenarios by selectively removing non-causal agents. This approach assumes that the removal of non-causal agents should not alter the ego agent's trajectory, thereby using the same ground truth trajectory to reinforce model predictions against changes under non-causal perturbations. We assess the performance in two key areas: in-distribution causal awareness and out-of-distribution generalization.

\begin{figure*}[t]
    \centering
    \begin{subfigure}[b]{0.45\linewidth}
        \centering
        \includegraphics[height=130pt]{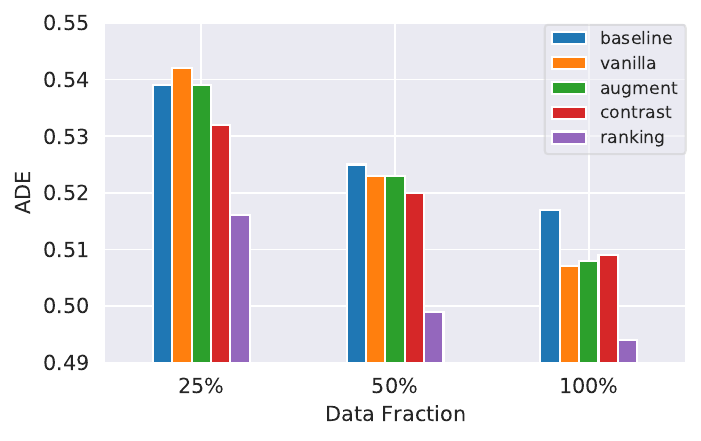}
    \end{subfigure}
    \quad
    \begin{subfigure}[b]{0.45\linewidth}
        \centering
        \includegraphics[height=130pt]{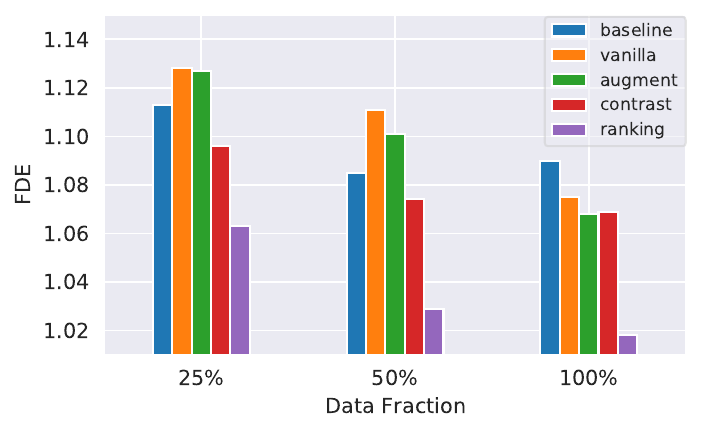}
    \end{subfigure}
    \vskip -0.1in
    \caption{\textbf{Results of causal transfer from ORCA simulations to the ETH-UCY dataset.}
    We consider three settings where the model has access to the same simulation data but varying amounts of real-world data.
    Our ranking-based causal transfer results in lower prediction errors and higher learning efficiency than the other counterparts, \ie, AutoBots baseline~\citep{girgisLatentVariableSequential2021}, the vanilla sim-to-real transfer~\citep{liangSimAugLearningRobust2020} and the non-causal data augmentation~\citep{roelofsCausalAgentsRobustnessBenchmark2022a}. Details are summarized in~\cref{appendix:implementation,appendix:results}.}
    \label{fig:ethucy}
    \vskip -0.1in
\end{figure*}

\subsubsection{In-distribution Causal Awareness}
\label{subsec:iid}

{\bf Setup.}
We first examine the efficacy of our method by measuring the ACE on the in-distribution test set.
Since the error of the baseline model varies substantially across different ranges of causal effects (\cref{subsec:recentrepresentation}), we split the test set into four quantile segments. The result is summarized in~\cref{fig:quantitative}.

{\bf Takeaway: our method boosts causal awareness thanks to fine-grained annotations.}
As shown in~\cref{fig:quantitative_ace}, both the contrastive and ranking variants of our causal regularization approach result in lower causal errors than the vanilla baseline.
In particular, the causal ranking method demonstrates substantial advantages over the other counterparts across all quantities of causal effects, confirming the promise of incorporating fine-grained annotations in learning causally-aware representations.

\subsubsection{Out-of-Distribution Generalization}
\label{subsec:ood}

{\bf Setup.}
We further examine the efficacy of our method by measuring the prediction accuracy on out-of-distribution test sets.
We consider two common types of distribution shifts in the multi-agent setting: higher density and unseen context, as detailed in~\cref{appendix:implementation}.
We summarize the results of different methods on the OOD test sets in~\cref{fig:quantitative_ood} and visualize the difference of prediction output in~\cref{fig:qualitative}.

\noindent{\bf Takeaway: causally-aware representations lead to stronger out-of-distribution generalization.}
As shown in~\cref{fig:quantitative_ood}, the prediction error from our causal ranking method is generally lower than that of the other methods on the OOD sets. In fact, the overall patterns between the two parts of ~\cref{fig:quantitative_ace} are highly similar, indicating a substantial correlation between causal awareness and out-of-distribution robustness. This practical benefit is also visually evident in the qualitative visualization in~\cref{fig:qualitative}.

\subsection{Effectiveness of Causal Transfer}
\label{subsec:s2r}

{\bf Setup.} Finally, we evaluate the proposed causal transfer approach on the ETH-UCY dataset~\citep{lernerCrowdsExample2007,pellegriniImprovingDataAssociation2010} and NBA rebound dataset \cite{xu2022socialvae,yue2014learning} paired with our diagnostic dataset. 
To assess the potential of the transfer method in challenging settings such as low-data regimes, we train Autobots~\citep{girgisLatentVariableSequential2021} on different fractions of the real-world data. The results are summarized in~\cref{fig:ethucy} and \cref{tab:nba}.

\noindent{\bf Takeaway: causal transfer improves performance in the real world despite domain gaps.} 
As shown in ~\cref{fig:ethucy}, our causal transfer approach consistently improves the performance of the learned representation on ETH-UCY, even with a substantial simulator-to-real-world gap.
In particular, it enables the model to learn faster, \eg, even with half data in the real-world, our ranking-based method can still outperform other counterparts.
In contrast, the vanilla sim-to-real technique~\citep{liangSimAugLearningRobust2020} which combines synthetic and real data, struggles with effective transfer, especially in low-data regimes.
We hypothesize this is due to disparities in non-causal interaction styles, such as speed and curvature, between simulation and reality. We report ORCA simulator and other baselines' prediction accuracy on ETH-UCY in~\cref{appendix:results}. We further examine the causal transfer approach on NBA rebound, a more large-scale and interactive dataset. Table \ref{tab:nba} shows that our ranking-based method outperforms the baseline and demonstrates that the proposed causal transfer can work not only for pedestrians but also for basketball players. 

\begin{table}[!tbp]
\centering
\begin{tabular}{lll}
\toprule
         & \multicolumn{1}{c}{ADE$\downarrow$}   & FDE$\downarrow$   \\ \midrule
Baseline                     & 0.562                     & 1.271 \\
Vanilla                      & 0.567                     & 1.287 \\
Augment                      & 0.569                    & 1.292 \\
Contrast (ours)               & 0.550                   & 1.248     \\
Ranking (ours)                     & \textbf{0.544}            & \textbf{1.235} \\ \bottomrule
\end{tabular}
\caption{\textbf{Results of causal transfer from ORCA simulations to the NBA rebound dataset.}We outperform the AutoBots baseline by integrating our causal regularizers, showing effectiveness of our approach. The numbers are in meters.}
\label{tab:nba}
\vskip -0.15in
\end{table}

\section{Conclusions}
\label{sec:conclusion}
{\bf Summary.}
In this paper, we presented a thorough analysis and an effective approach for learning causally-aware representations of multi-agent interactions.
We revisited the notion of non-causal robustness in the recent benchmark~\citep{roelofsCausalAgentsRobustnessBenchmark2022a} and showed that the main weaknesses of recent representations are not overestimations of the effect of non-causal agents but rather underestimation of the effects of indirect causal agents.
To boost causal awareness of the learned representations, we introduced a regularization approach that encapsulates a contrastive and a ranking variant leveraging annotations of different granularity.
We showed that our approach enables recent models to learn faster, generalize better, as well as transfer stronger to practical problems, even without real-world annotations.

\noindent{\bf Discussions.} As one of the first steps towards causally-aware representations in the multi-agent context, our work raises two primary questions for further exploration. 
On the technical front, while our proposed regularization method consistently boosts causal awareness in various settings, supervision alone is likely insufficient to fully solve causal reasoning in complex scenes, as evidenced by the causal errors in \cref{fig:quantitative_ace}. Incorporating structural inductive biases might be a promising direction to address high-order reasoning challenges in the presence of indirect causal effects.

\clearpage



\section{Acknowledgement}
\label{sec:acknowledgement}
The authors would like to thank Yang Gao and Kaouther Messaoud for their valuable feedback.
This work was supported by Honda R\&D Co., Ltd and Hasler Foundation under the responsible AI program.
{
    \small
    \bibliographystyle{ieeenat_fullname}
    \bibliography{bibtex/social,bibtex/ood,bibtex/causal,bibtex/orca,bibtex/ssl,bibtex/sim2real,bibtex/robotics,bibtex/yuejiang}
}
\clearpage

\newpage
\appendix
\onecolumn
\begin{table}[th]
\centering
\small

\resizebox{\linewidth}{!}{%
    \begin{tabular}{l c c c c c c c c c c c}
    \toprule
    25\% & ETH & HOTEL & UNIV & ZARA1 & ZARA2 & AVG \\            
    \midrule
    AutoBots~\citep{girgisLatentVariableSequential2021} & 0.942/1.886 & 0.333/0.662 & 0.563/1.156 & 0.446/0.958 & 0.409/0.904 & 0.539/1.113\\
    + Vanilla~\citep{liangSimAugLearningRobust2020} & 0.956/1.893 & 0.352/0.698 & \textbf{0.537/1.143} & 0.439/0.969 & 0.426/0.935 & 0.542/1.128 \\
    + Augment~\citep{roelofsCausalAgentsRobustnessBenchmark2022a} & 0.942/1.867 & 0.359/0.722 & 0.544/1.152 & 0.441/0.971 & 0.409/0.922 & 0.539/1.127 \\
    + Contrast (ours) & 0.938/1.885 & 0.320/0.616 & 0.565/1.187 & 0.450/0.971 & 0.386/0.821 & 0.532/1.096 \\
    + Ranking (ours) & \textbf{0.906/1.814} & \textbf{0.307/0.582} & 0.556/1.157& \textbf{0.435/0.942} & \textbf{0.376/0.819} & \textbf{0.516/1.063} \\
    \bottomrule
    \end{tabular}        
}
\caption{
    Quantitative results of sim-to-real transfer from ORCA simulations to 25\% of the ETH-UCY dataset.
}
\label{tab:ethucy25}
\vskip -0.2in
\end{table}

\begin{table}[th]
\centering
\small
\resizebox{\linewidth}{!}{%
    \begin{tabular}{l c c c c c c c c c c c}
    \toprule
    50\% & ETH & HOTEL & UNIV & ZARA1 & ZARA2 & AVG \\            
    \midrule
    AutoBots~\citep{girgisLatentVariableSequential2021} & 0.940/1.883 & 0.326/0.612 & 0.566/1.197 & 0.434/0.945 & 0.358/0.787 & 0.525/1.085\\
    + Vanilla~\citep{liangSimAugLearningRobust2020} & 0.923/1.913 & 0.342/0.661 & \textbf{0.535/1.141} & 0.430/0.954 & 0.383/0.886 & 0.523/1.111 \\
    + Augment~\citep{roelofsCausalAgentsRobustnessBenchmark2022a} & 0.937/1.885 & 0.340/0.660 & 0.535/1.146 & 0.424/0.938 & 0.377/0.878 & 0.523/1.101 \\
    + Contrast (ours) & 0.935/1.870 & 0.344/0.667 & 0.554/1.148 & 0.422/0.913 & 0.346/0.772 & 0.520/1.074 \\
    + Ranking (ours) & \textbf{0.903/1.810} & \textbf{0.300/0.566} & 0.543/1.146& \textbf{0.420/0.911} & \textbf{0.331/0.725} & \textbf{0.499/1.029} \\
    \bottomrule
    \end{tabular}        
}
\caption{Quantitative results of sim-to-real transfer from ORCA simulations to 50\% of the ETH-UCY dataset.
}
\label{tab:ethucy50}
\vskip -0.2in
\end{table}

\begin{table}[th]
\centering
\small
\resizebox{\linewidth}{!}{%
    \begin{tabular}{l c c c c c c c c c c c}
    \toprule
    100\% & ETH & HOTEL & UNIV & ZARA1 & ZARA2 & AVG \\            
    \midrule
    AutoBots~\citep{girgisLatentVariableSequential2021}             & 0.938/1.916 & 0.334/0.678 & 0.550/1.152 & 0.420/0.916 & 0.343/0.787 & 0.517/1.090\\
    + Vanilla~\citep{liangSimAugLearningRobust2020}                 & 0.923/1.913 & 0.331/0.638 & 0.527/1.133 & 0.410/0.907 & 0.346/0.783 & 0.507/1.075 \\
    + Augment~\citep{roelofsCausalAgentsRobustnessBenchmark2022a}   & 0.938/1.878 & 0.332/0.635 & \textbf{0.518/1.141} & \textbf{0.403/0.890} & 0.348/0.796 & 0.508/1.068 \\
    + Contrast (ours)                                               & 0.930/1.896 & 0.321/0.605 & 0.538/1.154 & 0.412/0.899 & 0.345/0.791 & 0.509/1.069 \\
    + Ranking (ours) & \textbf{0.900/1.780} & \textbf{0.296/0.556} & 0.539/1.148& 0.408/0.889 & \textbf{0.325/0.715} & \textbf{0.494/1.018} \\
    \bottomrule
    \end{tabular}        
}
\caption{Quantitative results of sim-to-real transfer from ORCA simulations to 100\% of the ETH-UCY dataset.
}
\label{tab:ethucy100}
\vskip -0.2in
\end{table}

\section{Additional Results} \label{appendix:results}
\subsubsection{AutoBots on ETH-UCY}
In addition to the aggregated results presented in~\cref{fig:ethucy}, we summarize an in-depth breakdown of the sim-to-real transfer results in~\cref{tab:ethucy25,tab:ethucy50,tab:ethucy100}.
Across all evaluated settings, our ranking-based causal transfer consistently achieves superior prediction accuracy compared to the AutoBots baseline. Notably, it outpaces the standard sim-to-real method~\citep{liangSimAugLearningRobust2020} in four out of the five subsets, with the sole exception in the UNIV subset.
We conjecture that this exception might be attributed to the high similarity between the ORCA simulation and UNIV dataset.

\begin{figure}[th]
    \centering
    \begin{subfigure}[b]{0.32\linewidth}
        \centering
        \includegraphics[width=1.0\linewidth]{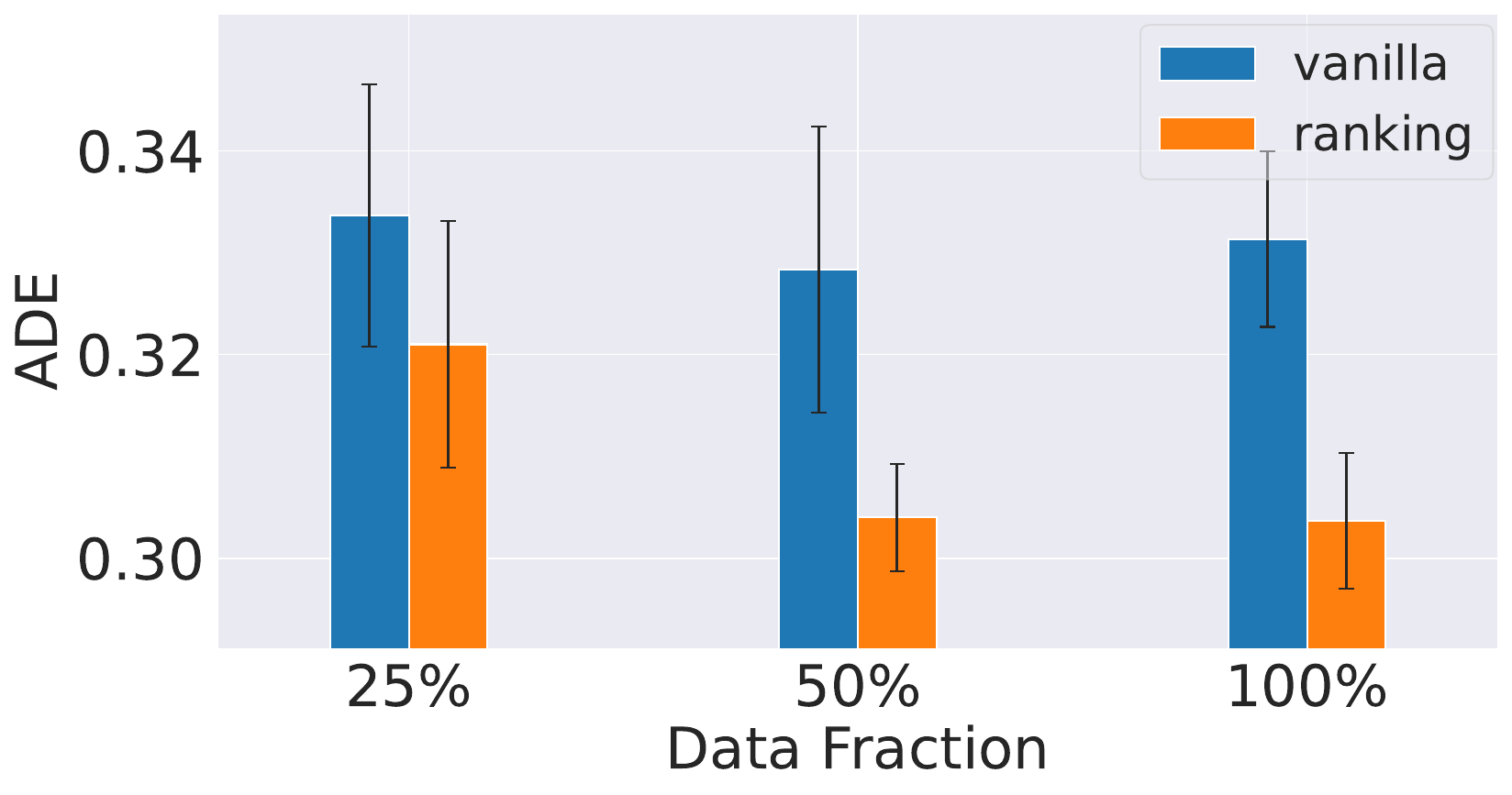}
        \caption{HOTEL}
        \label{fig:ade_hotel}
    \end{subfigure}
    ~
    \begin{subfigure}[b]{0.32\linewidth}
        \centering
        \includegraphics[width=1.0\linewidth]{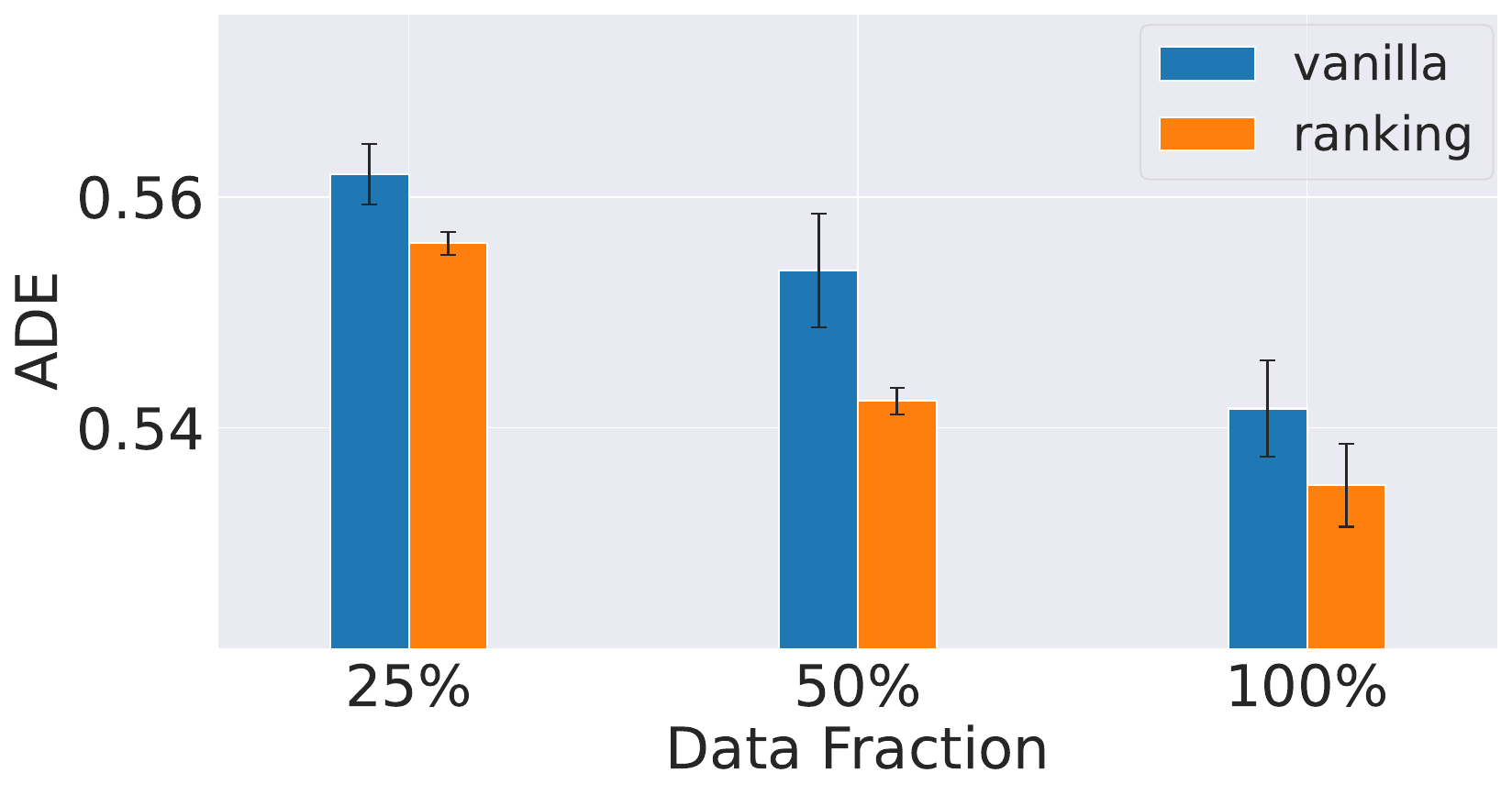}
        \caption{UNIV}
        \label{fig:ade_univ}
    \end{subfigure}
    ~
    \begin{subfigure}[b]{0.32\linewidth}
        \centering
        \includegraphics[width=1.0\linewidth]{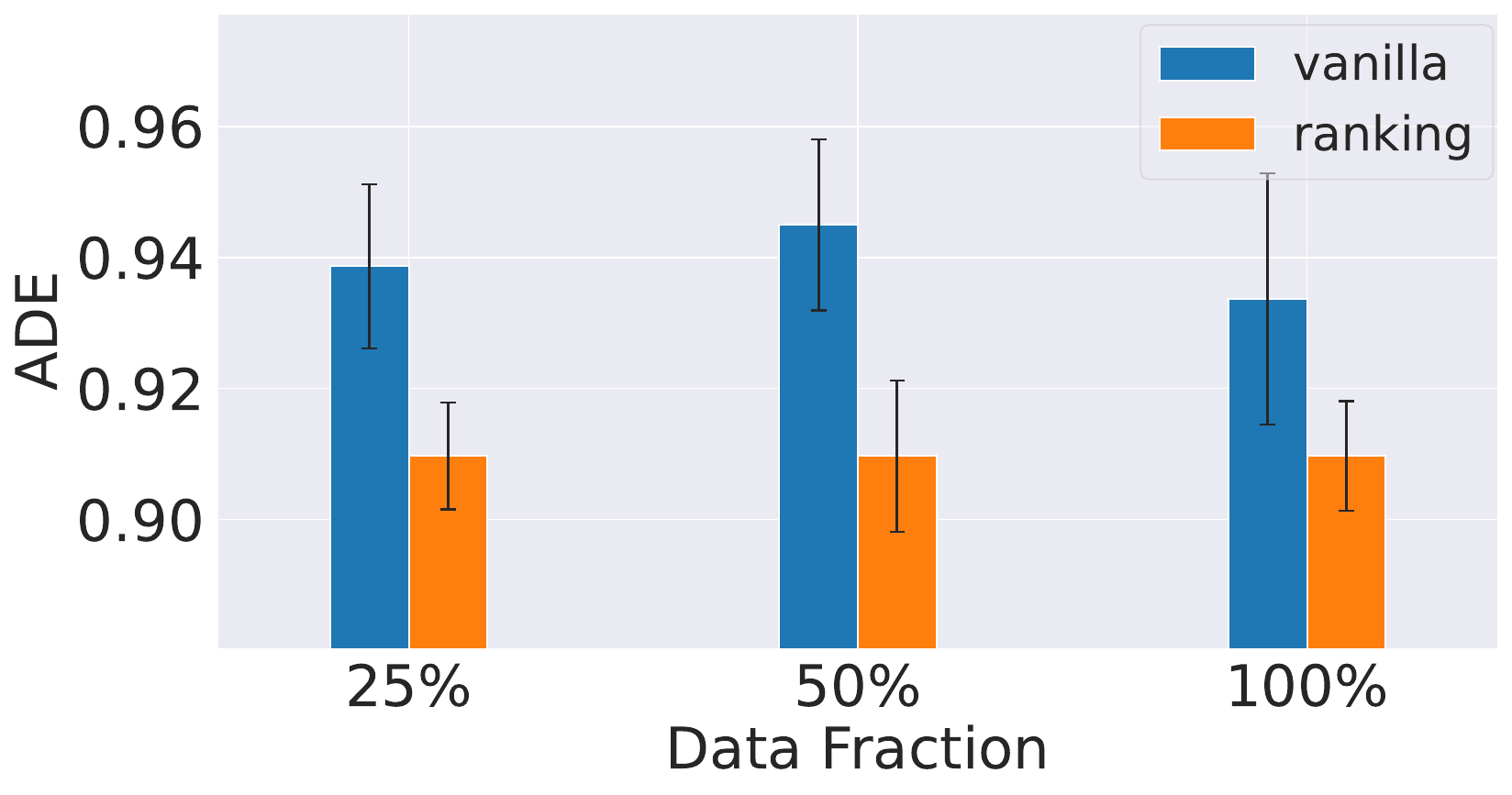}
        \caption{ETH}
        \label{fig:ade_eth}
    \end{subfigure}
    \begin{subfigure}[b]{0.32\linewidth}
        \centering
        \includegraphics[width=1.0\linewidth]{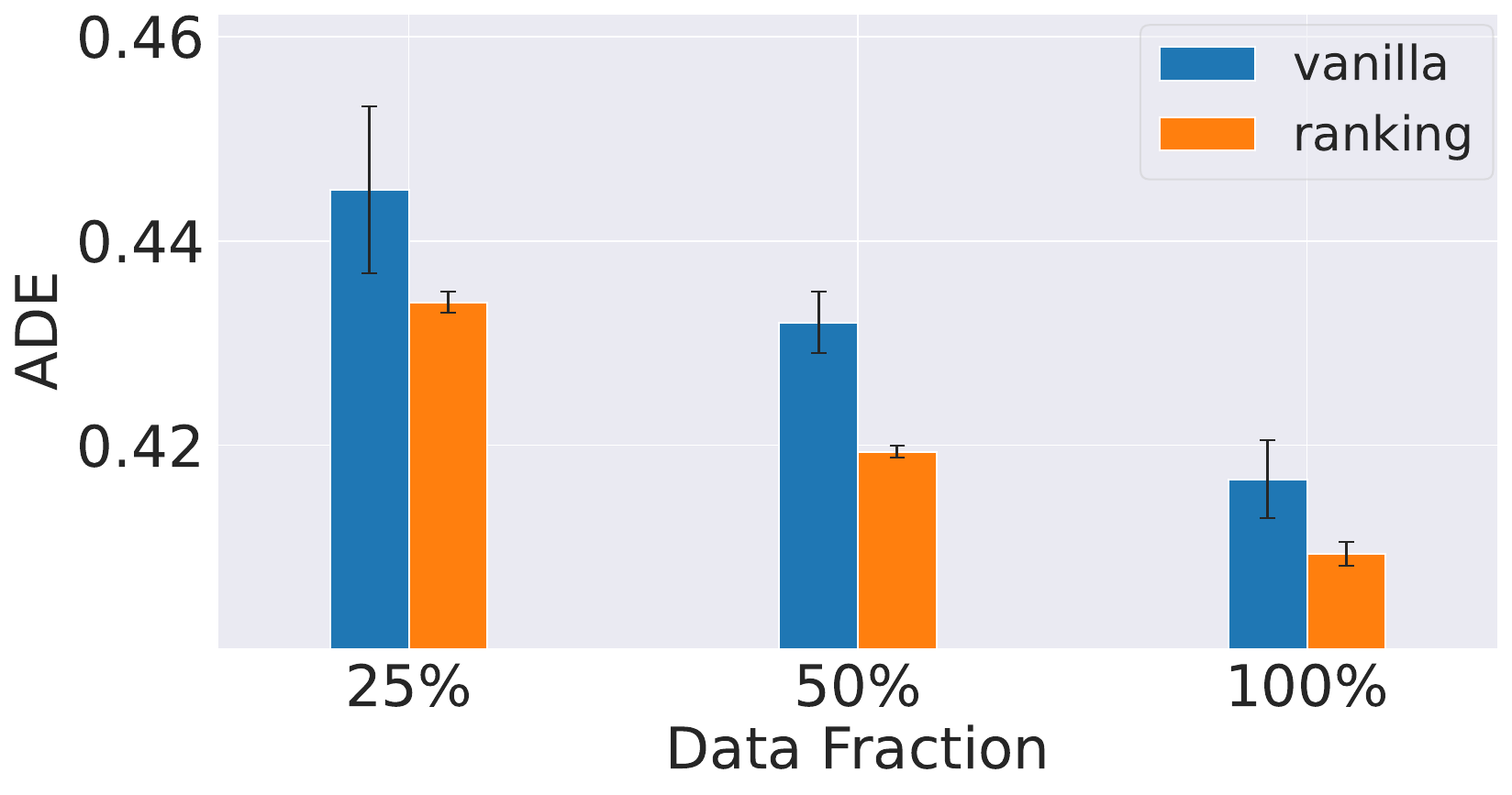}
        \caption{ZARA1}
        \label{fig:ade_zara1}
    \end{subfigure}
    ~
    \begin{subfigure}[b]{0.32\linewidth}
        \centering
        \includegraphics[width=1.0\linewidth]{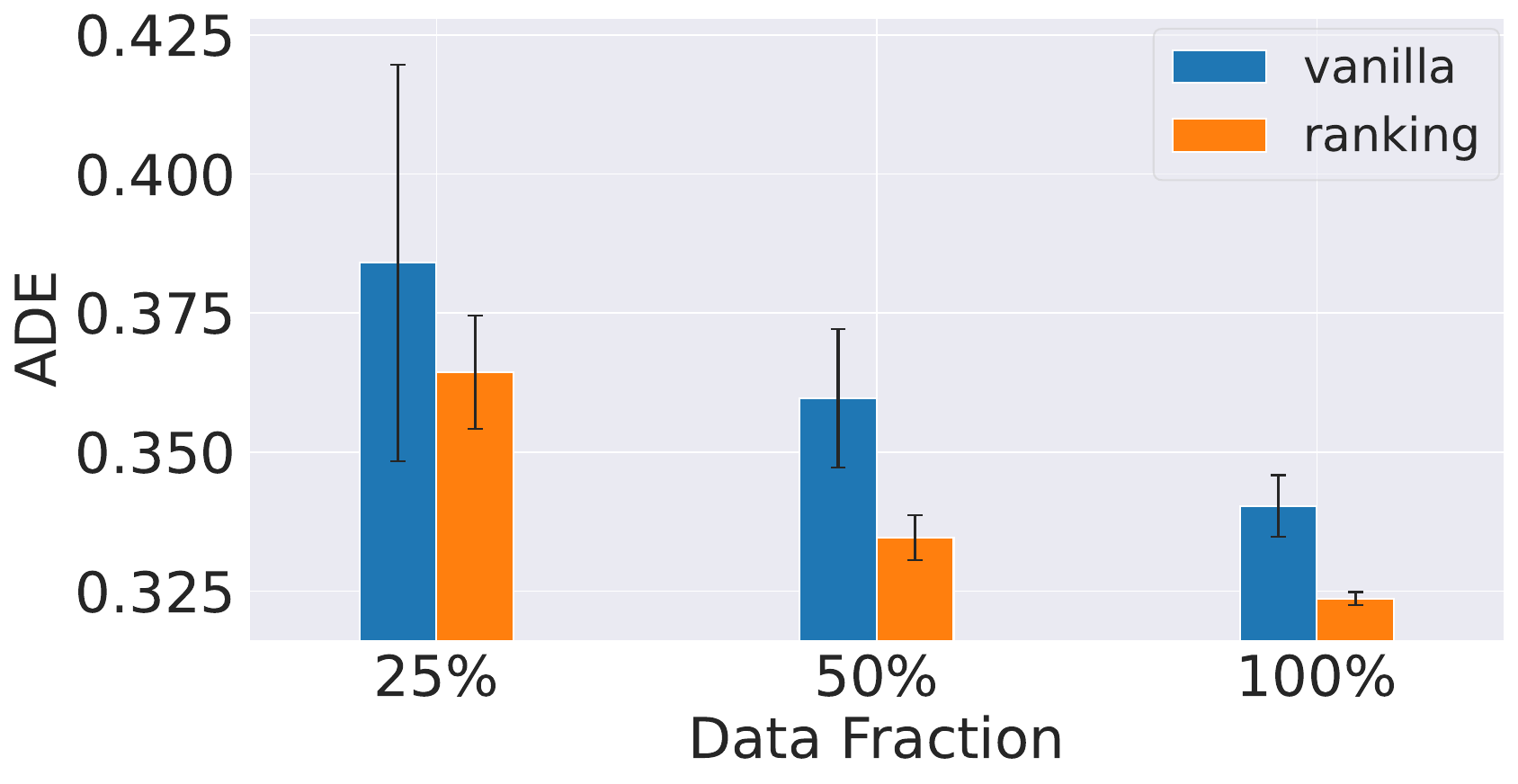}
        \caption{ZARA2}
        \label{fig:ade_zara2}
    \end{subfigure}
    ~
    \begin{subfigure}[b]{0.32\linewidth}
        \centering
        \includegraphics[width=1.0\linewidth]{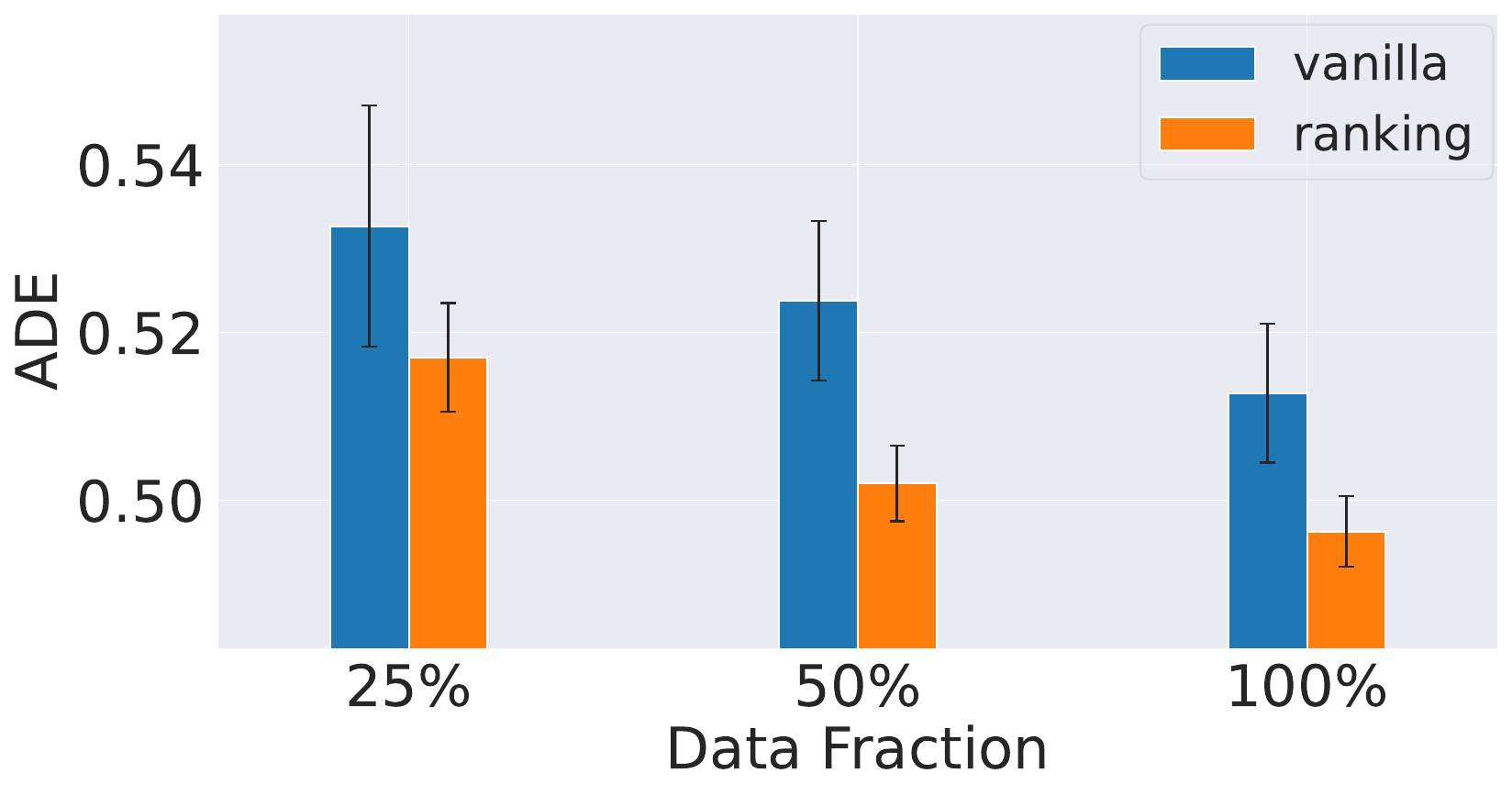}
        \caption{Average}
        \label{fig:ade_avg}
    \end{subfigure}

    \caption{Additional results of our ranking-based causal transfer on the ETH-UCY dataset, as a supplement to \cref{fig:ethucy}.
    The results of ADE are averaged on each subset over three random seeds.}
    \vskip -0.2in
    \label{fig:eth_ucy_ade}
\end{figure}

\begin{figure}[th]
    \centering
    \begin{subfigure}[b]{0.32\linewidth}
        \centering
        \includegraphics[width=1.0\linewidth]{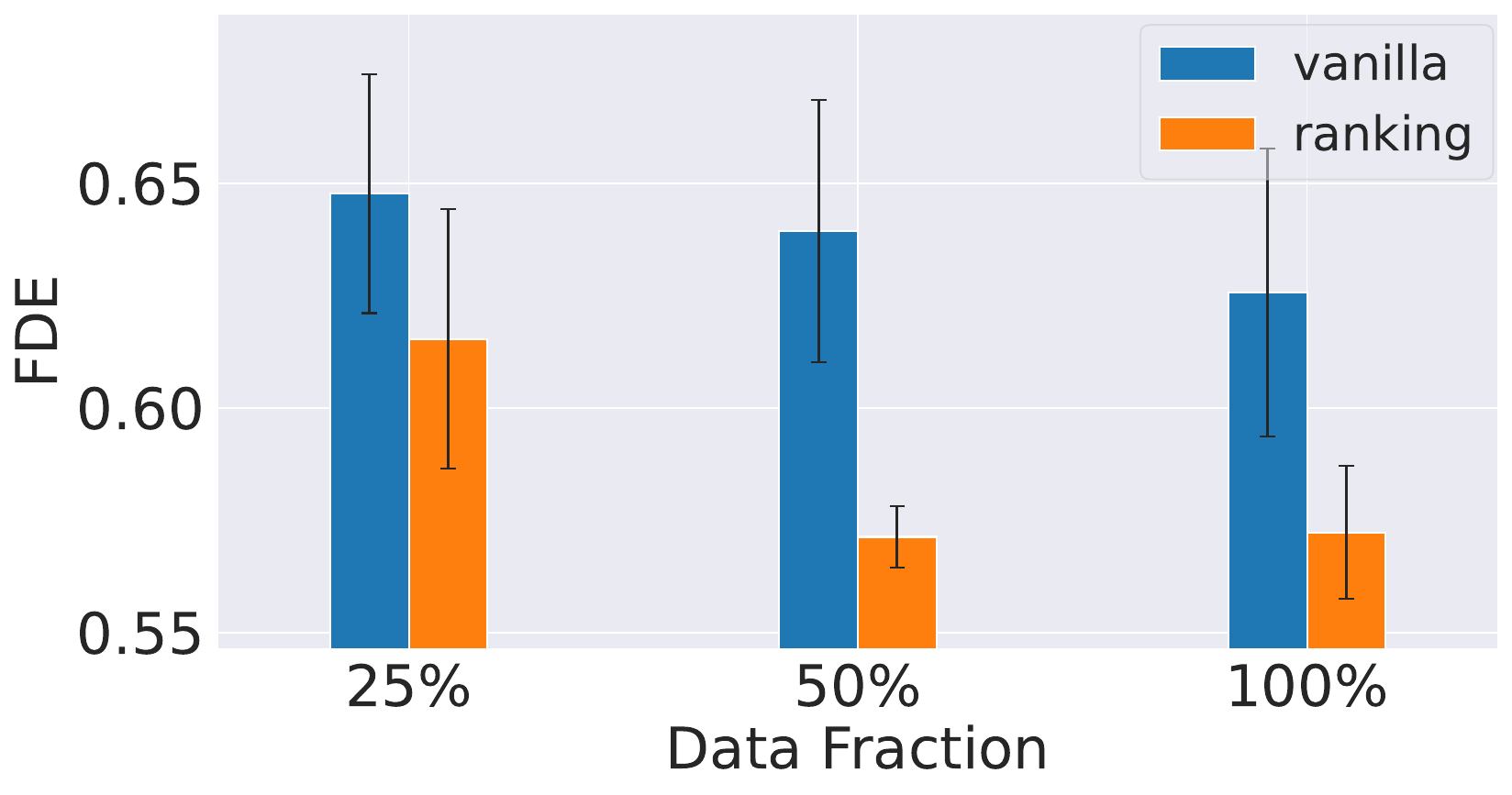}
        \caption{HOTEL}
        \label{fig:fde_hotel}
    \end{subfigure}
    ~
    \begin{subfigure}[b]{0.32\linewidth}
        \centering
        \includegraphics[width=1.0\linewidth]{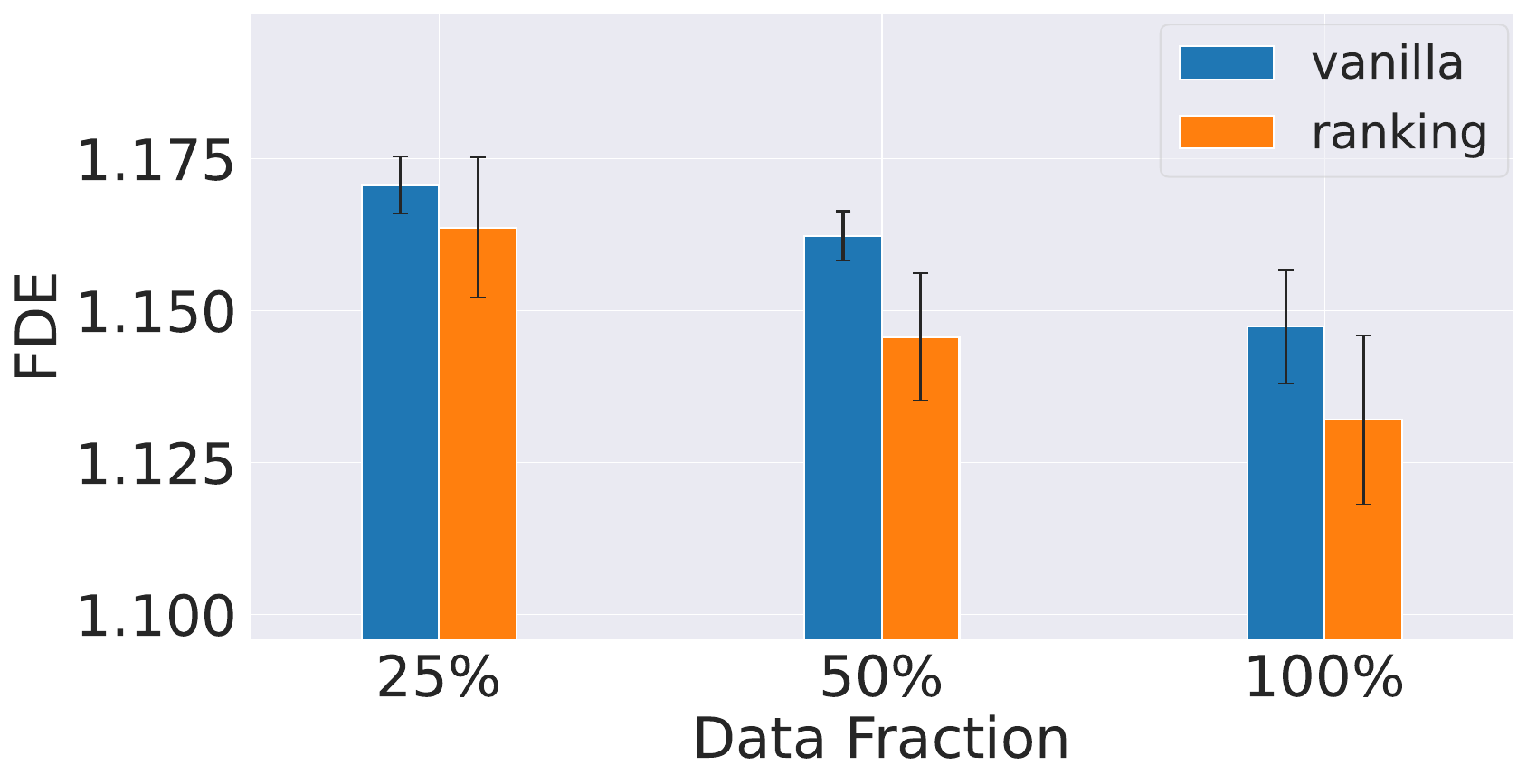}
        \caption{UNIV}
        \label{fig:fde_univ}
    \end{subfigure}
    ~
    \begin{subfigure}[b]{0.32\linewidth}
        \centering
        \includegraphics[width=1.0\linewidth]{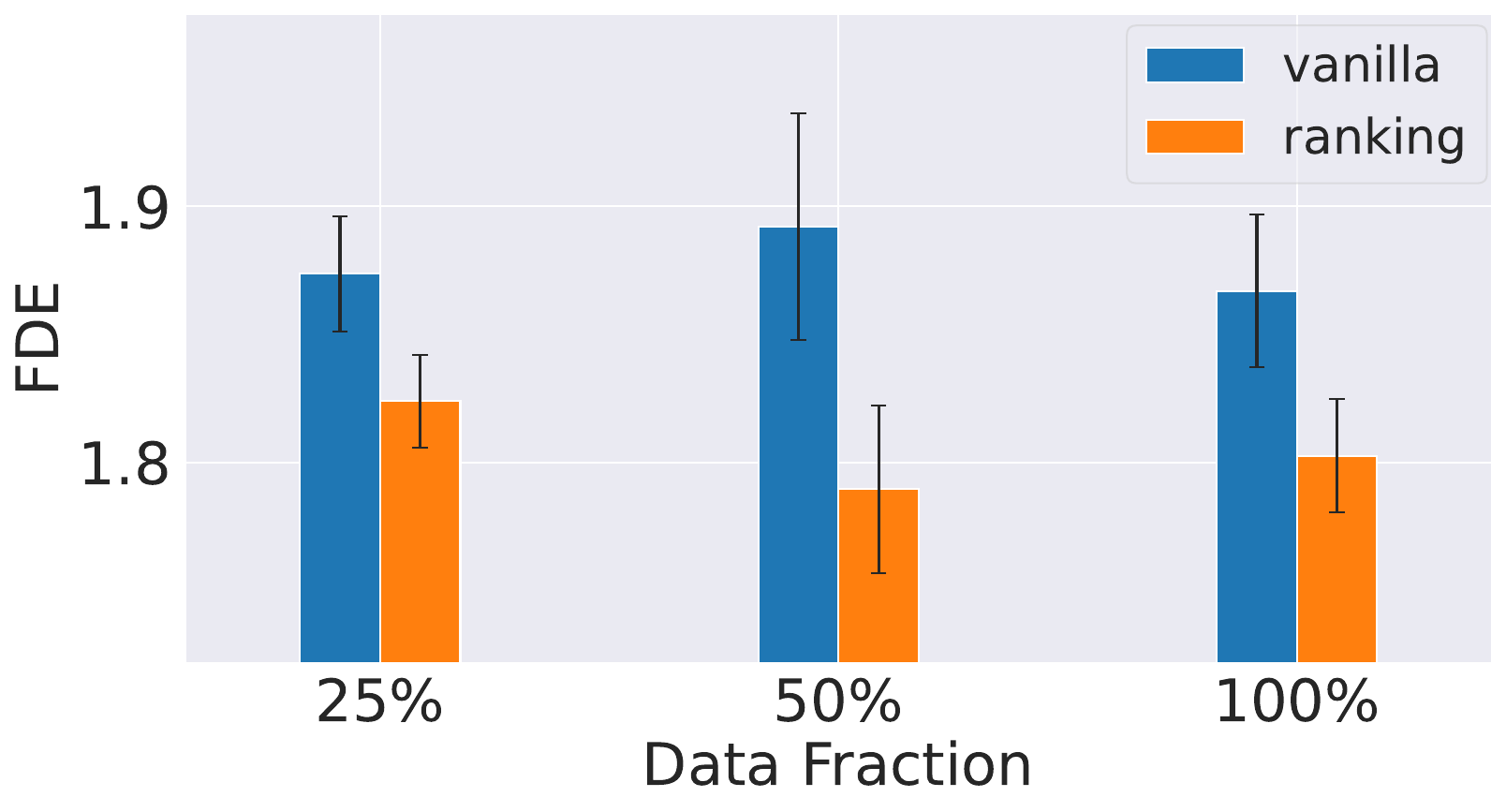}
        \caption{ETH}
        \label{fig:fde_eth}
    \end{subfigure}

    \begin{subfigure}[b]{0.32\linewidth}
        \centering
        \includegraphics[width=1.0\linewidth]{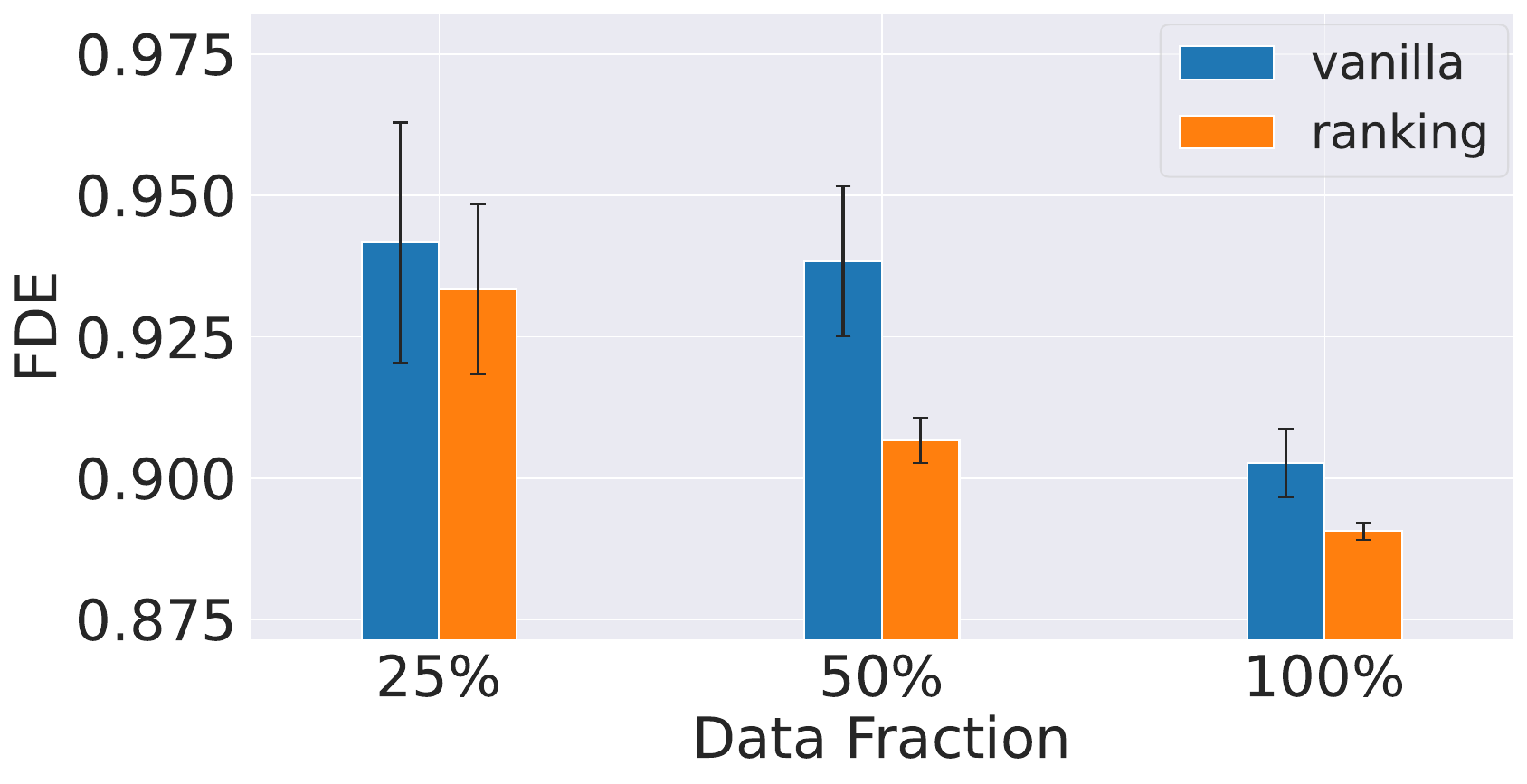}
        \caption{ZARA1}
        \label{fig:fde_zara1}
    \end{subfigure}
    ~
    \begin{subfigure}[b]{0.32\linewidth}
        \centering
        \includegraphics[width=1.0\linewidth]{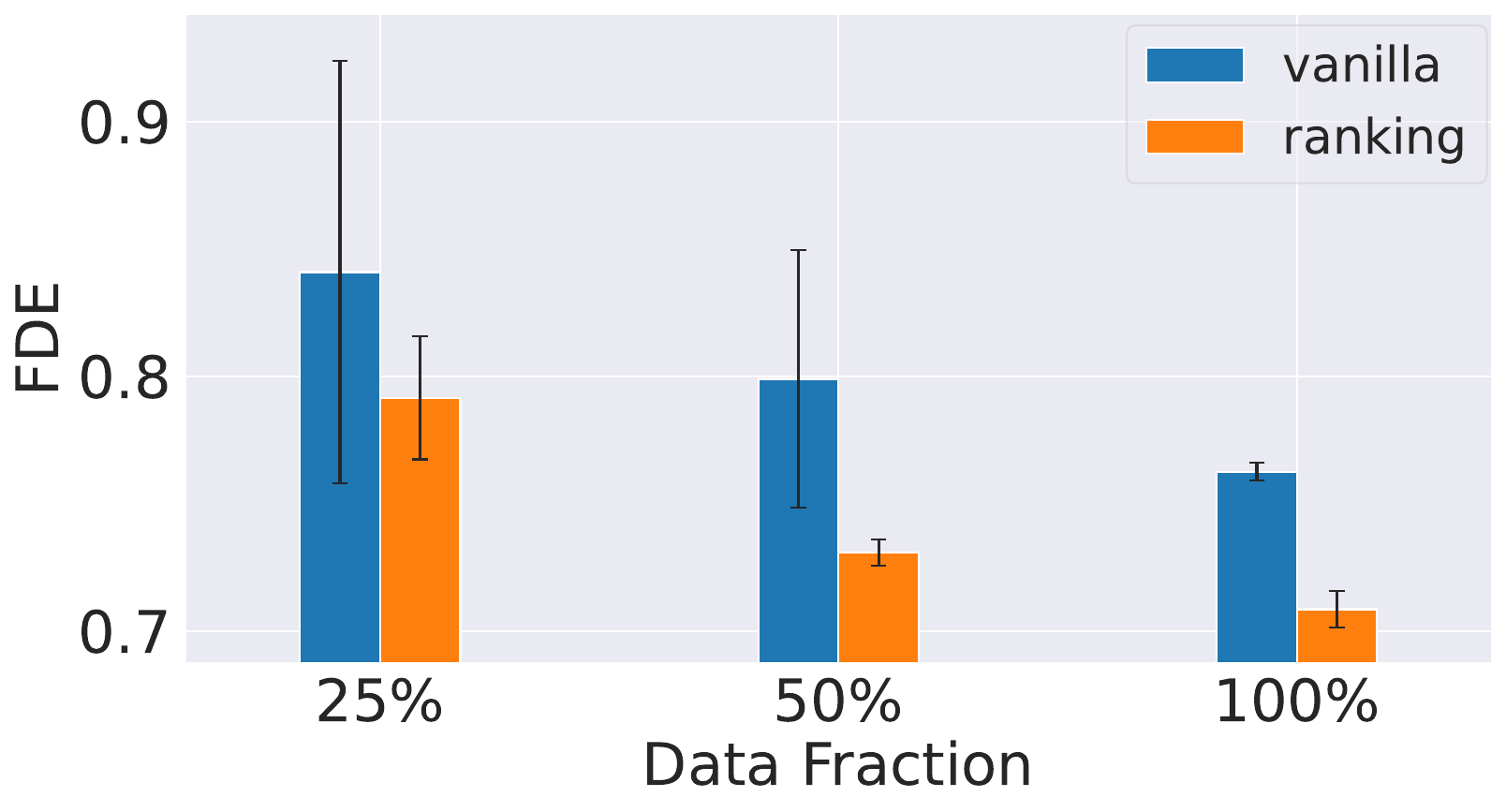}
        \caption{ZARA2}
        \label{fig:fde_zara2}
    \end{subfigure}
    ~
    \begin{subfigure}[b]{0.32\linewidth}
        \centering
        \includegraphics[width=1.0\linewidth]{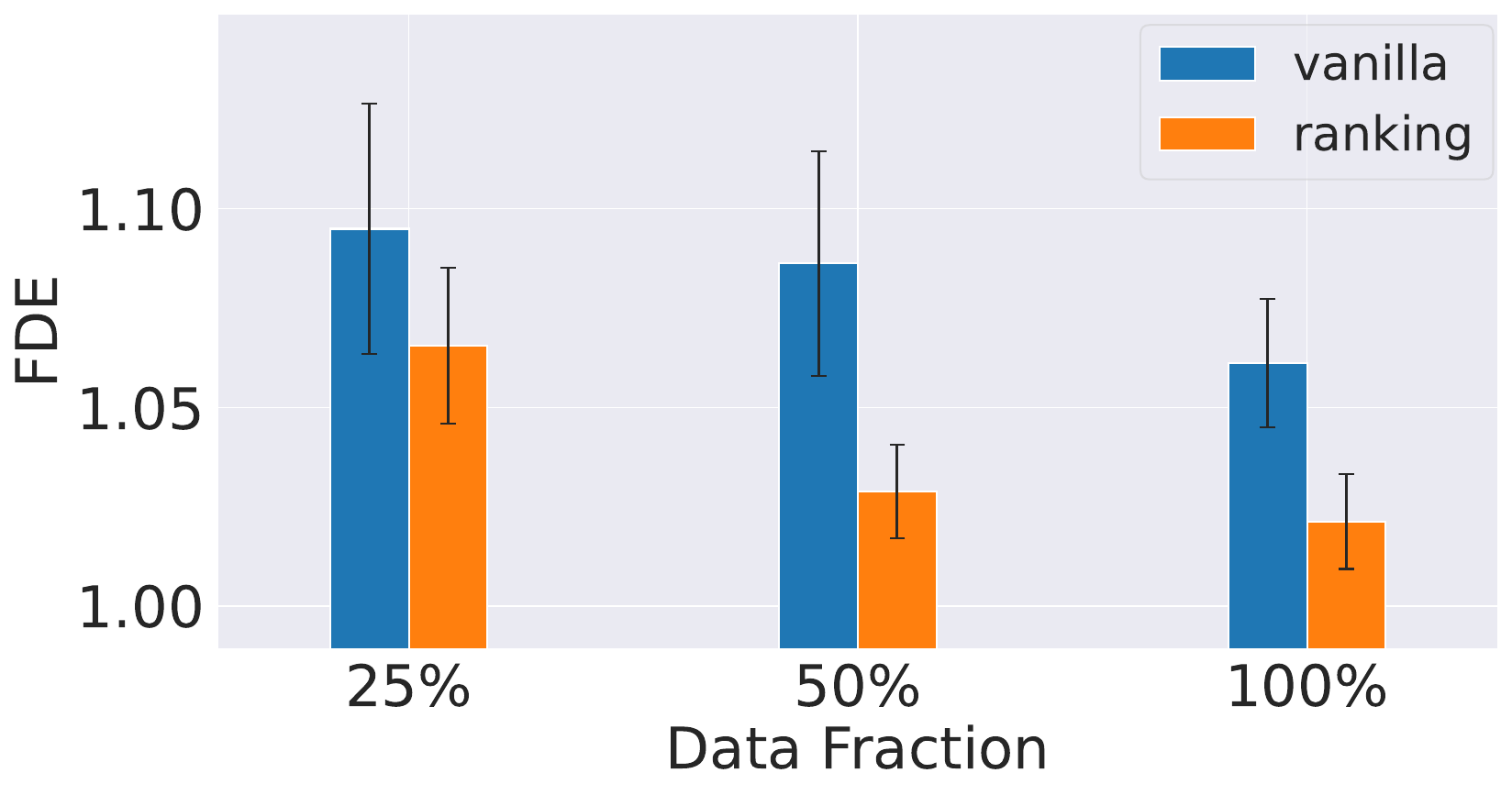}
        \caption{Average}
        \label{fig:fde_avg}
    \end{subfigure}

    \caption{Additional results of our ranking-based causal transfer on the ETH-UCY dataset, as a supplement to \cref{fig:ethucy}.
    The results of FDE are averaged on each subset over three random seeds.}
    \vskip -0.2in
    \label{fig:eth_ucy_fde}
\end{figure}

To assess the stability of the ranking-based method, we conducted experiments using three different random seeds, as illustrated in~\cref{fig:eth_ucy_ade,fig:eth_ucy_fde}. The results show that the ranking-based method consistently outperforms the vanilla model across all subsets of the ETH-UCY dataset. Furthermore, the ranking-based method typically yields smaller standard deviations, indicating more consistent performance.

Furthermore, as summarized in~\cref{tab:ethucysota}, AutoBots stands as one of the state-of-the-art models for multi-agent trajectory forecasting, leaving only marginal room for improvement on the ETH-UCY dataset.
In spite of this, our proposed causal transfer method still offers notable improvements, resulting in more enhanced predictions. Please also note that this improvement is achieved through ORCA, while ORCA's performance on ETH-UCY dataset is very poor, as shown in~\cref{tab:ethucysota}. This marks the effectiveness of our methodology in extracting and transferring causal knowledge from a simulator, despite its big gap to the real world data.

\begin{figure}[t]
    \vskip 0.2in
    \centering
    \begin{subfigure}[b]{0.375\linewidth}
        \centering
        \includegraphics[height=120pt]{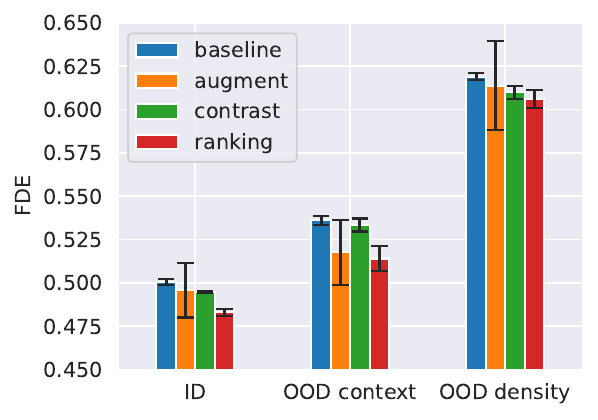}
    \end{subfigure}
    \caption{Additional quantitative results of our method on the out-of-distribution test sets, as a supplement to \cref{fig:quantitative_ood}. Models trained by our method yield lower FDE. Results are averaged over five random seeds.}
  \label{fig:OOD_FDE}
\end{figure}

Finally, we summarize the results of different methods in terms of 
Final Displacement Error (FDE) on the OOD test sets in~\cref{fig:OOD_FDE}. Similar to ~\cref{fig:quantitative_ood}, our ranking-based method leads to the lowest prediction errors compared to the other counterparts, reaffirming its strength for boosting out-of-distribution robustness.

\begin{table}[t]
\centering
\small
\resizebox{\linewidth}{!}{%
    \begin{tabular}{l c c c c c c c c c c c}
    \toprule
    Deterministic & ETH & HOTEL & UNIV & ZARA1 & ZARA2 & AVG \\            
    \midrule
    ORCA~\citep{vandenbergReciprocalVelocityObstacles2008}
    & 2.27/3.44 & 1.03/1.54  & 1.29/2.079 & 0.97/1.60  & 0.87/1.45 & 1.28/2.02 \\
    S-LSTM~\citep{alahiSocialLSTMHuman2016} & 1.09/2.35 & 0.79/1.76 & 0.67/1.40 & 0.47/1.00 & 0.56/1.17 & 0.72/1.54\\
    D-LSTM~\citep{kothariHumanTrajectoryForecasting2021} & 1.05/2.10 & 0.46/0.93 & 0.57/1.25 & 0.40/0.90 & 0.37/0.89 & 0.57/1.21 \\
    Trajectron++~\citep{salzmannTrajectronDynamicallyFeasibleTrajectory2020a} & 1.02/2.00 & 0.33/0.62 & 0.53/1.19 & 0.44/0.99 & 0.32/0.73 & 0.53/1.11 \\
    Social-Transmotion\citep{saadatnejad2024socialtransmotion} & 0.93/1.81 &  0.32/0.60 & 0.54/1.16 & 0.42/0.90 & 0.32/0.70 & 0.51/1.03 \\
    AutoBots ~\citep{girgisLatentVariableSequential2021} & 0.93/1.87 & 0.32/0.65 & 0.54/1.15 & 0.42/0.91 & 0.34/0.77 & 0.51/1.07 \\
    AutoBots + Ranking (ours) & \textbf{0.90/1.78} & \textbf{0.30/0.56} & 0.53/1.12 & 0.41/0.89 & 0.32/0.70 & \textbf{0.49/1.01} \\
    \bottomrule
    \end{tabular}        
}
\caption{Comparison between different multi-agent forecasting models on the ETH-UCY dataset. 
Boosted by our proposed ranking-based causal transfer, the best result of AutoBots across three random seeds reaches comparable performance to the current state-of-the-art.}
\label{tab:ethucysota}
\end{table}

\subsubsection{{Comparison with other robust representations}}
\begin{table}[!tbp]
\centering
\begin{tabular}{lll}
\toprule
     Dataset    & \multicolumn{1}{c}{ETH-UCY}   & NBA   \\ \midrule
AutoBots + Intervention \cite{chen2021human} & 0.53                   & 0.58 \\
AutoBots + Ranking (ours)  & \textbf{0.49}            & \textbf{0.54} \\ \bottomrule
\end{tabular}
\caption{\textbf{Comparison with causal intervention using AutoBot.} The ADE on ETH-UCY and NBA verify our causal transfer is robust under domain shifts across different datasets.}
\label{tab:intervention}
\end{table}
We compare our ranking regularizer with causal intervention \cite{chenHumanTrajectoryPrediction2021}. The results, detailed in \cref{tab:intervention}, demonstrate that our method effectively mitigates the domain shift between simulated and real environments and better enhances performance compared to the proposed causal intervention. 

\subsubsection{{Muti-Transmotion on JRDB}}
\begin{table}[!tbp]
\centering
\begin{tabular}{lll}
\toprule
     Model    & \multicolumn{1}{c}{K=1}   & K=5   \\ \midrule
Multi-Transmotion                    & 0.379                    & 0.272 \\
Multi-Transmotion + Ranking (ours)  & \textbf{0.342}            & \textbf{0.264} \\ \bottomrule
\end{tabular}
\caption{\textbf{ADE of causal transfer from ORCA simulations to the JRDB dataset.} We outperform the Muti-Transmotion baseline by integrating our causal regularizers, showing the effectiveness of our approach in deterministic (K=1) and multimodal (K=5) setups. The numbers are in meters.}
\label{tab:jrdb}
\end{table}

We further verify our causal transfer on a new model and dataset. We adopt Multi-Transmotion \cite{gao2024multi}, a SOTA human motion prediction model, on JRDB\cite{martin2021jrdb}, a social navigation dataset. We split an indoor and an outdoor scene as the testing set, and the other as the training set. The results in \cref{tab:jrdb} validate the effectiveness of our causal regularize, which consistently improves the model, to demonstrate the proposed ranking regularizer can be implemented in various baseline scene representations and has the potential to improve them in many settings. In addition, the second column in \cref{tab:jrdb} shares our improvement in the multimodal setup. In this multi-task approach, we learn the multimodal trajectory prediction task on the real-world dataset, while at the same time we regularize the representation learned by the encoder using our causal ranking regularizer, on the synthetic data. 
\section{Implementation Details} \label{appendix:implementation}

\paragraph{Experiment details.}

Our experiments are largely built upon the public code of prior work, with as few modifications as possible made for the implementations of our proposed regularizers.
Concretely, in the robustness analysis reported in \cref{tab:baselines}, we train each model on our constructed dataset using the default hyperparameters of the corresponding baseline.
To understand the performance of our proposed methods in \cref{subsec:synthetic}, we fine-tune the pre-trained checkpoint for 10 epochs, and evaluate the obtained model on the hold-out test set. The main hyperparameters used for training our baseline model AutoBots~\citep{girgisLatentVariableSequential2021} and the causal regularizers are listed in \cref{tab:hyper}.

\paragraph{Simulation details.}

Our diagnostic dataset is generated using a customized version of the Reciprocal Velocity Obstacle simulator that employs Optimal Reciprocal Collision Avoidance (ORCA) \citep{berg2011reciprocal}.
To simulate realistic causal relationships between agents, we imposed a visibility constraint where an agent only observes other neighbors within their proximity and its \ang{210} field of view. This visibility plays a significant role in determining the influence of one agent on another. Specifically, we define a neighbor $i$ as having a {\it direct influence} on the ego agent at time step $t$ if it is visible to the ego in that time step, \ie, $\mathbbm{1}^i_t=1$.
Additionally, we introduce a visibility window that records agents that were previously visible, facilitating the modeling of a richer spectrum of direct and indirect inter-agent influences.
To encourage the presence of non-causal agents in dense spaces, we explicitly directed specific agents to follow others, thereby making them non-causal or indirect causal.

\paragraph{Dataset details.}

\cref{tab:dataset} summarize the key statistics of our diagnostic datasets, including both the in-distribution (ID) training set and the out-of-distribution (OOD) test set.
Specifically, we consider two distinct types of OOD datasets, each deviating from the ID dataset in specific aspects, such as agent density and/or scene context.
\begin{itemize}[nosep]
    \item {\em ID}: The training dataset is characterized by an average of 12 pedestrians per scene, each interacting with a few others to navigate towards their goals. All the scenes are set in an open area context, allowing unrestricted movements and serving as the base environment in our experiments.
    \item {\em OOD Density}: In our first OOD set, we retain the same context setting as the ID dataset but increase agent density. Specifically, we introduce more agents in proximity to the ego agent to intensify agent interactions. Additionally, we add agents behind the ego agent, which results in more non-causal agents. This dataset aims to test the robustness of the model in handling increased agent density.
    \item {\em OOD Context}: The second OOD set alters the scene context from an open area to a narrow street, where pedestrians walk from one end to the other. Given that agents walking in the same direction generally do not interact, we double the number of agents in the scene, thus ensuring a similar degree of interaction complexity to the ID dataset.
\end{itemize}
Exemplary animations for each data split can be found in our public repository.


\paragraph{Baseline details.}
To the best of our knowledge, few prior work studies causally-aware representation learning in the multi-agent context.
To examine the efficacy of our proposed methods, we consider the following three existing methods as comparative baselines.
\begin{itemize}[nosep]
    \item {\em Baseline}: the baseline method trains the model on the data in the target domain only, \ie, the AutoBots baseline trained on the in-distribution dataset in \cref{subsec:synthetic} or the real-world ETH-UCY dataset in \cref{subsec:s2r}.
    \item {\em Vanilla}: the vanilla sim-to-real method combines simulated and real-world data in the training process. It adheres to a standard prediction task, with an equal mix of data from each domain in every training batch~\citep{liangSimAugLearningRobust2020}. 
    \item {\em Augment}: the causal augmentation method is built upon the {\em Baseline} for the experiment in \cref{subsec:synthetic} or the {\em Vanilla} for the experiment in \cref{subsec:s2r}. It augments training data by randomly dropping non-causal agents based on the provided annotations~\citep{roelofsCausalAgentsRobustnessBenchmark2022a}.
\end{itemize}

\begin{figure}[t]
    \centering
    \begin{subfigure}[b]{0.45\linewidth}
        \centering
        \includegraphics[height=100pt]{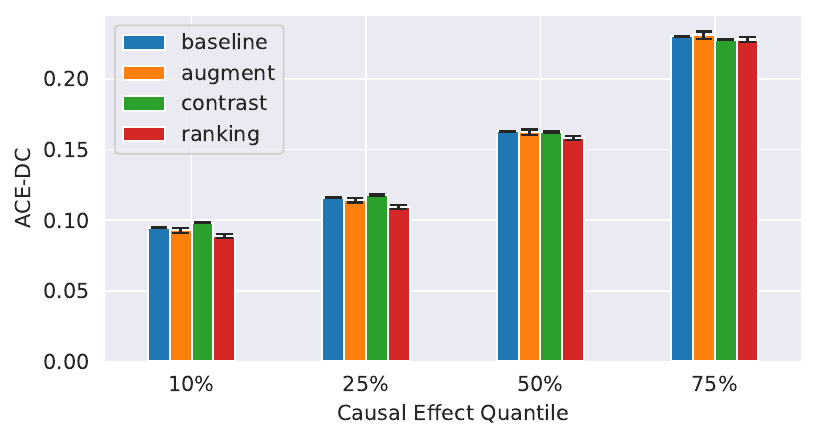}
    \end{subfigure}
    \quad
    \begin{subfigure}[b]{0.45\linewidth}
        \centering
        \includegraphics[height=100pt]{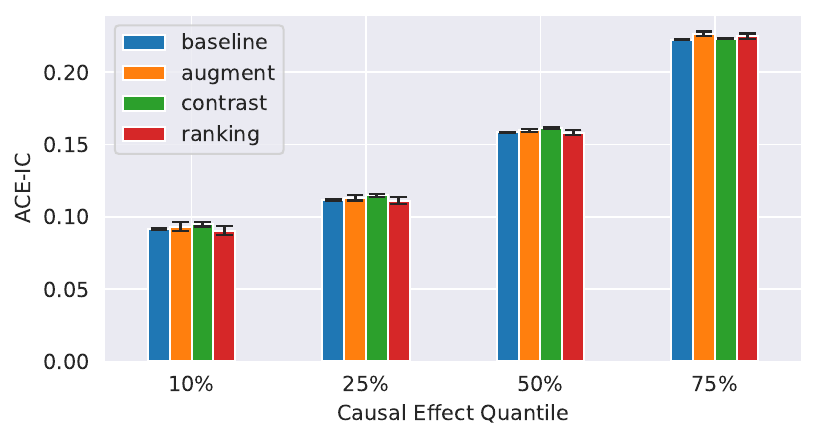}
    \end{subfigure}
    \caption{Additional quantitative results of our method on causal effect estimates, as a supplement to \cref{fig:quantitative_ace}.
    Models trained by our method yield lower ACE-DC and ACE-ID, especially noticeable in scenarios with low-quantile ground-truth causal effects. Results are averaged over five random seeds.}.
    \label{fig:ace}
\end{figure}


\section{Additional Discussions} \label{appendix:discussion}

\paragraph{Counterfactual simulation.}
Our diagnostic dataset, enabled by counterfactual simulations, offers clean annotations of causal relationships, serving as a crucial step in understanding causally-aware representation of multi-agent interactions.
However, the realism of these simulated causal effects is still subject to some inherent limitations.
For example, we have enforced a stringent constraint on the field of view for each agent, considering that the ego agent is usually unaffected by trailing neighbors. Such constraints could compromise the optimality of the ORCA algorithm, potentially resulting in unnatural trajectories.
We believe that integrating more advanced simulators, \eg, CausalCity~\citep{mcduffCausalCityComplexSimulations2022}, can address these challenges and we anticipate promising outcomes along this line for future research.

\paragraph{Multi-agent causal effects.} \label{appendix:caveat_eval}
Our annotation and evaluation have been focused on the causal effect at an individual agent level, namely we remove only one agent at a time.
This aligns with the notion of Causal Agents~\citep{roelofsCausalAgentsRobustnessBenchmark2022a}, \ie, a neighboring agent has a certain causal relationship with the ego agent.
Through this lens, we observe that while recent representations are already partially robust to non-causal agent removal, they tend to underestimate the effects of causal agents.
However, it is important to note that this is still a rather simplified and restricted setting compared to the group-level causal effects, where the collective behavior of multiple agents may have a more complex influence on the ego agent. 
Understanding and addressing this challenge can be another exciting avenue for future research.


\begin{table}[th]
\centering
\begin{minipage}[b]{0.41\textwidth} 
\centering
\small
    \begin{tabular}{wl{5.2cm}|wl{1.6cm}}
        name & value \\        
    \specialrule{.1em}{.05em}{.05em}  
        batch size & 16 \\
        pre-training learning rate &  $7.5\times10^{-4}$\\
        fine-tuning learning rate &  $2.34375\times10^{-5}$\\
        contrastive weight $\alpha$ & 1000 \\
        ranking weight $\alpha$ & 1000 \\
        ranking margin $m$ & 0.001 \\
        non-causal threshold $\epsilon$ & 0.02 \\
        causal threshold $\eta$ & 0.1 \\   
    \end{tabular}
    \caption{
    Key hyper-parameters in our experiments.
    }
    \label{tab:hyper}
\end{minipage}
\hfill
\begin{minipage}[b]{0.55\textwidth} 
\centering
\resizebox{\textwidth}{!}{%
    \begin{tabular}{l | c c | c c c c}
    \toprule
     \multirow{2}{*}{Dataset} & \multicolumn{2}{c}{Number of scenes} & \multicolumn{4}{c}{Number of agents per scene} \\ 
 &   train  & test & non-causal & direct causal & indirect causal & total \\
 \midrule
     ID & 20k  & 2k & 1.31 & 8.35 & 0.48 & 13.03\\  
     OOD Context & - & 2k & 6.13 & 9.92 & 1.47 & 21.39\\
     OOD Density & - & 2k & 9.47 & 12.27 & 2.55 & 29.98\\
    \bottomrule
    \end{tabular}        
}
\caption{Key statistics of our diagnostic datasets.}
\label{tab:dataset}
\end{minipage}%

\end{table}

\section{Qualitative Examples of Synthetic Data}
\begin{figure*}[t]
    \centering
    \begin{subfigure}[b]{0.47\linewidth}
        \centering
        \includegraphics[width=1.0\linewidth]{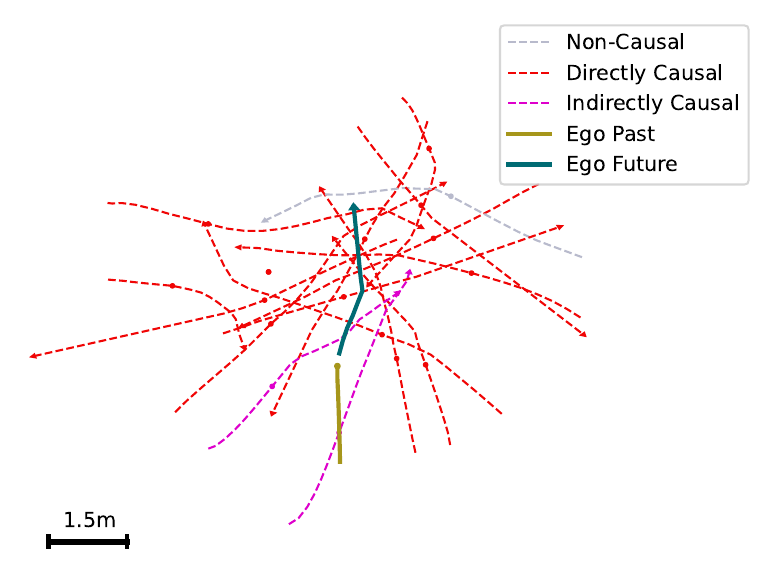}
    \end{subfigure}
    ~
    \begin{subfigure}[b]{0.47\linewidth}
        \centering
        \includegraphics[width=1.0\linewidth]{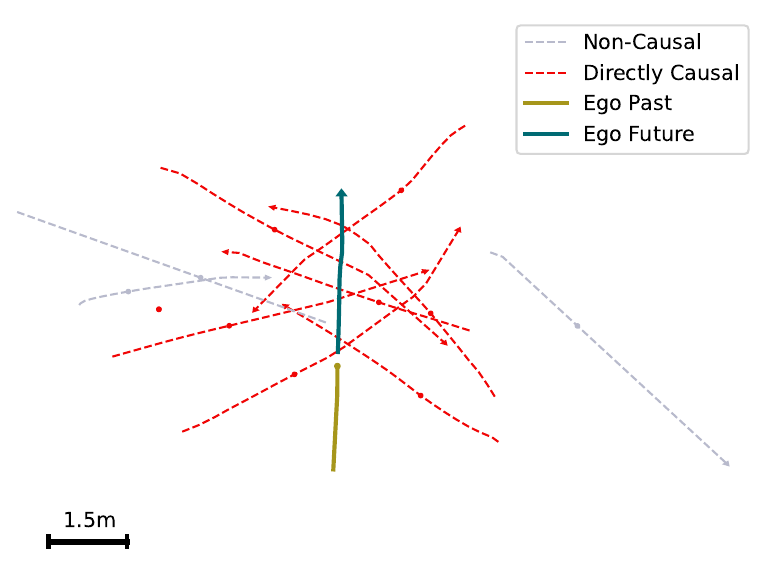}
    \end{subfigure}
    ~
    \begin{subfigure}[b]{0.47\linewidth}
        \centering
        \includegraphics[width=1.0\linewidth]{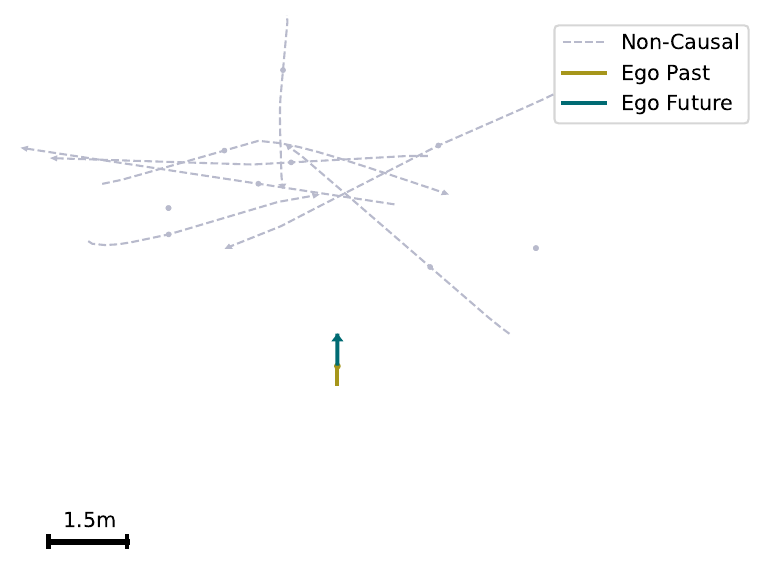}
    \end{subfigure}
    ~
    \begin{subfigure}[b]{0.47\linewidth}
        \centering
        \includegraphics[width=1.0\linewidth]{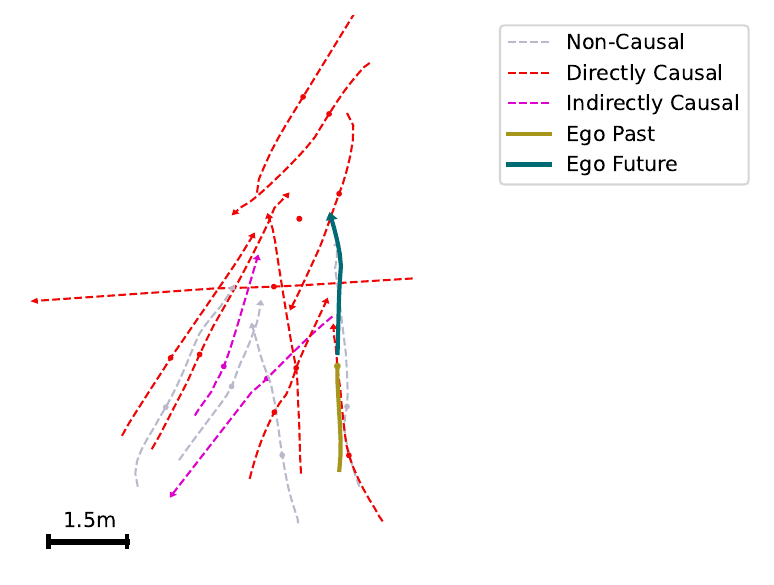}
    \end{subfigure}
    ~
    \begin{subfigure}[b]{0.47\linewidth}
        \centering
        \includegraphics[width=1.0\linewidth]{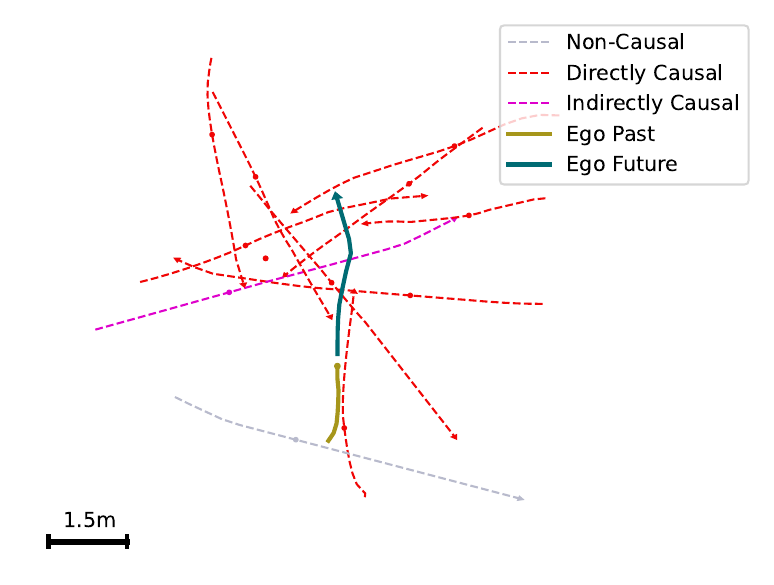}
    \end{subfigure}
    ~
    \begin{subfigure}[b]{0.47\linewidth}
        \centering
        \includegraphics[width=1.0\linewidth]{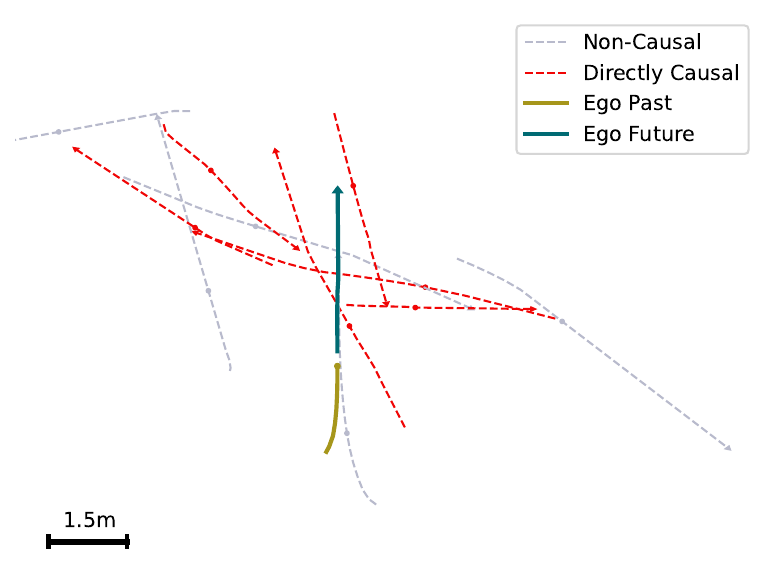}
    \end{subfigure}
    ~
    \caption{\textbf{Visualizations of our synthetic dataset with causal labels, generated using ORCA simulator.} The dots indicate the position of the agent in the last timestep before the prediction horizon, and single dots without any lines associated with them represent static agents who do not move in the scene.}
    \label{fig:qualitative-data}
\end{figure*}

In this section we share some qualitative examples of our synthetic dataset, with the non-causal, directly causal, and indirect causal labels in them. Some scenarios are depicted in \cref{fig:qualitative-data}, where the points on the lines represent the position of agents in the last time step before the prediction horizon. Lonely dots without any lines on them represent static agents that do not move in the scene. 

\end{document}